\documentclass[10pt,journal,compsoc]{IEEEtran}

\ifCLASSOPTIONcompsoc
  \usepackage[nocompress]{cite}
\else
  \usepackage{cite}
\fi

\usepackage{bbding}
\usepackage[misc]{ifsym}
\usepackage{graphicx}
\usepackage{float}
\usepackage{color}
\usepackage[ruled]{algorithm2e}
\usepackage{caption}
\usepackage{subfig}
  \usepackage[switch]{lineno}
\usepackage{url}
\usepackage{booktabs}

\ifCLASSINFOpdf
\else
\fi
\usepackage{color}
\usepackage[cmex10]{amsmath}
\usepackage{bbding}
\usepackage[misc]{ifsym}
\usepackage{graphicx}
\usepackage{float}
\usepackage{color}
\usepackage[ruled]{algorithm2e}
\usepackage{subfig}
\usepackage{amsmath,amssymb}
\usepackage{multirow}
\usepackage{mathrsfs}
\usepackage{dsfont}
\usepackage{bm}

\newtheorem{definition}{Definition}

\newtheorem{thm}{\bf Theorem}[section]
\newtheorem{remark}{Remark}
\newtheorem{pro}{\bf Proposition}

\newcommand{\dx}{{\mathrm{d} x}}

\begin{document}
\title{Deep Convolutional Neural Networks with Spatial Regularization, Volume and Star-shape Priori for Image Segmentation}
%
%
%

\author{Jun~Liu, Xiangyue~Wang, Xue-cheng~Tai
\IEEEcompsocitemizethanks{\IEEEcompsocthanksitem 
Jun Liu, Xiangyue Wang are with School of Mathematical Sciences, Laboratory of Mathematics and
Complex Systems, Beijing Normal University, Beijing 100875, P.R. China.\protect}
\IEEEcompsocitemizethanks{\IEEEcompsocthanksitem Xue-cheng Tai is with the Department of Mathematics, Hong Kong Baptist University, Kowloon Tong, Hong Kong.}
}

\IEEEtitleabstractindextext{%

\begin{abstract}
We use Deep Convolutional Neural Networks (DCNNs)
for image segmentation problems. DCNNs can well extract the features from natural images. However, the
classification functions in the existing network architecture of CNNs are simple and lack capabilities to handle  important spatial information in a way that have been done for many well-known traditional variational models. Prior such as spatial regularity, volume prior and object shapes cannot be well handled by existing DCNNs.
We propose a novel Soft Threshold Dynamics (STD) framework which can easily integrate many spatial priors of the
classical variational models into the DCNNs for image segmentation. The novelty of our method is to interpret the softmax activation function as a dual variable in a variational problem, and thus many spatial priors can be imposed in the dual space. From this viewpoint,
 we can build a STD based framework which can enable the outputs of  DCNNs to have many special priors such as spatial regularity, volume constraints and star-shape priori. The proposed method is a general mathematical framework and it can be applied to any semantic segmentation DCNNs.
 To show the efficiency and accuracy of our method, we applied  it to the popular DeepLabV3+ image segmentation network, and the experiments results show that our method can work efficiently on data-driven image segmentation DCNNs.
\end{abstract}

}

\maketitle
\IEEEdisplaynontitleabstractindextext
\IEEEpeerreviewmaketitle

%

%
\IEEEpeerreviewmaketitle

\IEEEraisesectionheading{\section{Introduction}}
\IEEEPARstart{I}mage segmentation is a fundamental task in the field of computer vision.
It is an important branch in the field of Artificial Intelligence (AI) and machine vision technology. It is of great significance in the fields of automatic driving, indoor navigation, medical image diagnosis, wearable equipment, virtual reality, and augmented reality \emph{etc.}.

 For traditional variational 
models, image segmentation is to divide an image into several disjoint regions at the pixel levels by given a single image. The model-based image segmentation methods mainly use some pre-given  prior information set by the models. They can combine the gray levels, color, spatial textures, geometry and some other features of the image pixels as a  similarity measure term. To enforce spatial priori such as spatial regularization, volume and shape information, this kind of model often contains another term called regularization term, which can make the solutions (segmentations) of the model  belong to a proper function space. Usually, the segmentation is carried out by extracting the determined low-level features of images.
Therefore, the segmentation method based on handcraft model is convenient to integrate various spatial prior information such as spatial regularization, sparsity, volume constraint, shape priori \emph{etc.} into the segmentation algorithms. Many outstanding research results have been done in this field.

In model-based image segmentation, the variational method has achieved great success because of its easy modeling process, simple process of implementation and a large number of mature algorithms. In general, due to the use  of the regularization term, the original ill-posed problem can be transformed into a well-posed problem. 
Among them, the famous total variation (TV) \cite{chan2001active} regularization has been widely used for its high segmentation accuracy. It is well-known that TV is not smooth and not easy to solve though many fast algorithms \cite{Wang2008,Goldstein2009,huang2008fast} have been developed. To achieve the similar regularization effects, but with better properties, an alternative method is the Threshold Dynamics (TD) method\cite{merriman1992diffusion,MBO2,EsedoThreshold}. In applications to image segmentation, TD  may have slightly less accuracy than TV, but its  computational efficiency is much higher. The TD method was developed by Merriman, Bence, and Osher \cite{merriman1992diffusion}\cite{MBO2} (also called MBO scheme) for the motion of an interface driven by the mean curvature.
It can converge to a continuous motion by mean curvature \cite{Evans1993Convergence}. This method was extended to the multiphase image segmentation in \cite{Tai2007} and multiphase flow with arbitrary surface tension in \cite{esedoglu2015threshold}, respectively. It attracted people's attention by its simplicity and unconditional stability. Subsequently, it has been extended to many other applications, including image processing \cite{TD2006}, \cite{Wang2017},\cite{MerkurjevAn}, interface motion problem with area or volume preservation \cite{Ruuth2003A}, anisotropic interface motion problem \cite{MerrimanConvolution}, graph cut and data clustering \cite{vanGennip2014}, auction dynamics \cite{jacobs2018auction}, \emph{etc.}.
In addition, many spatial prior information can be added to the segmentation models with this approach.

\begin{figure*}
\centering
\includegraphics[width=1.0\linewidth]{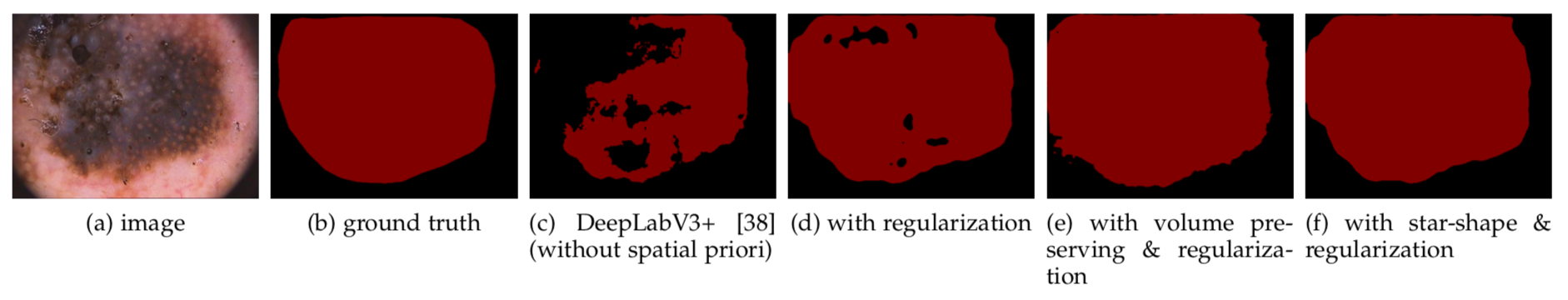}
\caption{An example of without and with the proposed spatial priori for DeepLabV3+ on ISIC2018 validation set.
The spatial priori such as regularization, volume preserving, star-shape in both forward and backward  propagations can improve the segmentation.
}\label{fig:show}
\end{figure*}

However, the model-based method lacks the ability of learning from the big data. When there are a large mount of samples, it is usually difficult to extract good deep features by handcraft designed models.

With the improvement of computing power, people began to consider the semantic image  segmentation by deep learning. This requires that the models can extract more advanced semantic features of images, which are difficult to be defined manually and subjectively. Recently, the data driven Deep Convolutional Neural Networks (DCNNs) method has been successfully applied in many computer vision and image processing problems due to its good feature extracting ability. In the image semantic segmentation field, there have been many DCNNs for image semantic segmentation.

With the emergence of Fully Convolutional Network (FCN) \cite{Long2014Fully}, deep learning  entered the field of image semantic segmentation. FCN enables DCNNs to carry out dense pixel classification prediction by  fusing multiscale and deep features. After that, this paradigm was adopted in
almost all the advanced approaches in the field of semantic segmentation  such as \cite{liu2015semantic,ghiasi2016laplacian,lin2016efficient,lin2017exploring}.

To better decompose the images into sparse components and reconstruct intrinsic features with multi-scale, two different architectures have been  proposed.
U-Net\cite{Ronneberger2015} puts forward a U-shaped Encoder-Decoder structure to solve this problem, it concatenates multi-scale features in the channel dimension to form thicker features. This Encoder-Decoder structure has been widely used and modified. In 2019, a SegNet \cite{SegNet} based on Encoder-Decoder structure adopted the method of recording the location of pooling and transferring the indices  to decoder, which reduced the model scale compared with U-Net and improved the segmentation resolution. U-Net has been greatly expanded in the field of medical image, such as Unet++\cite{zhou2018unet++}, V-net\cite{milletari2016v}, Progressive Dense V-net (PDV-Net)\cite{brosch2016deep}. Based on U-Net, many networks have been developed for other segmentation tasks \cite{cciccek20163d,zhang2018road}. The second architecture is to use the dilated/atrous convolution rather than the pooling layer. Dilated convolution\cite{YuMulti} plays an important role in maintaining the spatial resolution of the final feature map. But it brings heavy complexity and memory consumption at the same time. FastFCN\cite{FastFCN} modified it at the network structure level by proposing Joint Pyramid Sampling (JPU) block to replace the dilated convolution. There are many other models using multi-scale analysis for segmentation such as \cite{zhao2017pyramid,he2019dynamic,ding2018context,he2019adaptive,lin2018multi,li2017instance}.

DeepLabV3+\cite{v3+} combines the advantages of atrous spatial pyramid pooling (ASPP) module and Encoder-Decoder structure, and it has become the baseline of segmentation DCNNs  since  the first version of DeepLab model\cite{v1} was proposed six years ago. A series of continuous improvement and optimization for DeepLab on feature extractor, target scale modeling, context information processing, and model training process have been studied to upgrade it as the DeepLabV2\cite{v2} and DeepLabV3\cite{v3}. In 2018, DeepLabV3+\cite{v3+} has been expanded from DeepLabV3 by adding a simple and effective decoder module to refine segmentation results. The deep wise separable convolution to the ASPP and decoder modules is further applied  to obtain a faster and more powerful semantic segmentation network. The paper\cite{v3+} shows that this model achieves a high accuracy on Pascal VOC 2012 segmentation dataset.

We can see from the network structure in the literature that none of these mentioned network architectures contains spatial dependency information. 
To add the spatial smoothness information into DCNN,
there were several attempts.

The first method is a post-processing technique. The Conditional Random Field (CRF)\cite{crf} post-processing
belongs to this method. CRF is a graphic model of smooth segmentation based on image intensity, edges and other features. It can improve the segmentation accuracy, and was applied in DeepLabV1,V2,\cite{zheng2015conditional,arnab2016higher,monteiro2018conditional} \emph{etc.}. However, the spatial information can not be propagated in the training step because the post-processing structure does not join in the back propagation. The second method is to modify the loss function by adding loss with spatial information such as image edges. Brabandere et al.\cite{loss1} use discrimination loss based on distance measurement learning principle to consider spatial information. Liu et al.\cite{ERN} added the boundary information loss of the middle layer to the total loss function. This kind of method show its efficiency on improving the quality of the segmentation results since the spatial prior 
can be learned by the DCNN according to the back propagation of loss function.
But the loss function located after the prediction stage is usually dropped out when we apply a trained DCNN for prediction, and thus the spatial prior can not be well preserved in DCNN prediction procedure. 
Another potential problem of this method is that the DCNN is a good  fitting
function, and the loss functions do not play so important role in the training as 
model-based method. The third method is to adjust network structure. Based on data extraction and enhancement of attention gates,
Takikawa et al. \cite{TakikawaGated} proposed a two-stream DCNN architecture which coordinates a regular stream and a shape stream to combine the feature and boundary information. This structure has been shown that it can improve the segmentation results on small objects. At present, this attention mechanism \cite{li2019expectation,niu2020hmanet,huang2019ccnet,zhao2018psanet} has been widely studied. In very recent, to keep the spatial regularization into the DCNNs, TV regularization was integrated into the semantic image segmentation DCNNs in our previous work \cite{Jia2019}. To overcome the non-smooth of TV in the back propagation, the dual method is adopted. However, such a dual method often requires hundreds of iterations to reach convergence and its extremely high computational costs is a problem sometimes.

DCNNs can well extract abundant and high-level features that belong to the category itself, but it often cannot well keep the basic features that the image segmentation task requests. That is, some existing objective prior information,  such as small intraclass variance, piecewise constants segmentation \cite{MumfordOptimal}, smooth segmentation edges \cite{Potts1952}\cite{chan2001active}, \emph{etc.}
cannot be well learned by a general DCNN. This is because 
the existing DCNNs are just continuous mappings, and they fail to describe these complicated spatial priors.

Compared to image classification, another problem of DCNNs based semantic segmentation is the loss of location information caused by the pooling layers which are added to receive large receptive field for multi-scale. Thus, the spatial priori can help DCNNs to recover location information if we can add the known spatial priori.

On the other hand, it is a superiority for traditional model-based image segmentation to preserve the spatial regular features since the segmentation of the model can be well restricted on some specific function spaces.

To be different from the existing methods,
we will propose a framework to integrate the classical handcraft
model into the DCNN architecture. In our method, the spatial prior can be propagated in both forward and backward propagations.
What is more, our method can be easily extended to any  semantic image  segmentation DCNNs since it is a general mathematical framework which can combine prior information into DCNN architectures. As an application, we will take 
the popular semantic segmentation DeepLabV3+ as an example to show how it works. By applying our (variational) layers into the DeepLabV3+ basic network, we can make the spatial prior information to contribute to the training process  both in the forward and backward propagations in the network, and thus it can essentially improve the segmentation results. An example of with and without the proposed spatial priori on DeepLabV3+ is shown in \figurename\ref{fig:show}. As can be seen from this figure,
the segmentation results would be greatly improved if the images have a generic spatial priori.

The main contributions of this paper include:
\begin{itemize}
\item We proposed a general mathematical framework to enable the model-based image segmentation techniques to be be applied into constructions of image segmentation DCNN architectures.  Firstly,
to fit the model-based image segmentation method,
the activation functions in the DCNNs are reinterpreted as a 
minimizer of a variational problem, which enables us to force
the output of the DCNNs belong to specific function spaces and combine many spatial priors; Secondly, 
a Soft 
Threshold Dynamics (STD) method is proposed to integrate 
the spatial priors such as regularization, volume constraint and star-shape priori into DCNNs. In STD, the regularization term and entropic term are both smooth. When it is used with DCNNs, the forward and backward  propagations are both stable and  fast convergent.

\item A number of unconditional stability STD based algorithms and  related STD based blocks for DCNNs are proposed. These new DCNN blocks can keep the spatial priors such as boundary smoothness, volume preserving, and star-shape. It is difficult for traditional DCNNs to handle these important spatial prior. 

\item To show the simplicity with implementation and efficiency of the proposed method, the STD blocks based DeepLabV3+ for image segmentation are developed by taking 
the DeepLabV3+ as the basic network.  Our method combines the superiority of both the DCNNs and the model-based image segmentation methods. Experimental results show that our method can improve the accuracy of segmentation. It can improve many state-of-the-art image segmentation DCNNs.
\end{itemize}

The paper is organized as follows. In Section \ref{sec2}, we summarize the DCNN structure from a mathematical viewpoint and review the classical model-based image segmentation techniques. In Section \ref{sec3}, the proposed method is introduced.
This part includes:  the motivation of this paper; a variational explanation for softmax activation function; a STD method and the related stable algorithms with volume preserving and star-shape priors;  and STD based DCNN blocks together with their applications on DeeplabV3+.
In Section \ref{sec4}, we show  numerical results to verify the high efficiency of the proposed method. The conclusion will be given in Section \ref{sec5}.


\section{The related works}\label{sec2}
\subsection{DCNNs based image segmentation}
Let  $\bm v^{0}: \Omega\rightarrow\mathbb{R}$ be an input image of a pixel-wise segmentation neural network. The image segmentation network can be written as a parameterized nonlinear operator $\mathcal{N}_{\bm\Theta}$ defined by $\bm v^T=\mathcal{N}_{\bm\Theta}(\bm v^0)$. The output $\bm v^T$ of a network is given by the following $T$ layers recursive connections
\begin{equation}\label{eq:nn1}
\left\{
\begin{array}{rl}
\bm o^t=&\mathcal{T}_{\bm \Theta^{t-1}}(\bm v^{t-1},\bm v^{t-2},\cdots,\bm v^0),\\
\bm v^{t}=&\mathcal{A}^{t}(\bm o^t), t=1,\cdots, T.
\end{array}
\right.
\end{equation}
Here $\mathcal{A}^{t}$ is an activation functional such as sigmoid, softmax, ReLU \emph{etc.}. It also can be downsampling, upsampling operators and their compositions.
$\mathcal{T}_{\bm \Theta^{t-1}}(\bm v^{t-1},\bm v^{t-2},\cdots,\bm v^0)$ is a given operator which shows the 
connections between the $t$-th layer $\bm v^{t}$ and its previous layers $\bm v^{t-1},\bm v^{t-2},\cdots,\bm v^0$.
For the  simplest convolution network, $\bm v^{t}$ is usually only associate to
$\bm v^{t-1}$ and 
$\mathcal{T}_{\bm \Theta^{t-1}}(\bm v^{t-1})= \mathcal{\bm W}^{t-1} \bm v^{t-1} + \bm b^{t-1}$ is an affine transformation, in which $\mathcal{\bm W}^{t-1}, \bm b^{t-1}$ are linear operator (e.g. convolution) and translation, respectively.
$\bm\Theta=\{\bm\Theta^t=(\mathcal{\bm W}^{t}, \bm b^t)| t=0,T-1\}$ is an unknown parameter set.
The output of this network $\bm v^T:\Omega\rightarrow [0,1]^{I}$ should a soft classification function (e.g. softmax)
whose component function $v^T_{i}(x)$ implies the probability of a pixel located at $x$  belongs to $i$-th class.

To extract the multi-scale features of images, a lot of network  architectures have been proposed.
FCN is a successful end-to-end convolutional network for image  semantic segmentation.
By carefully choosing the operators $\mathcal{A}^{t}$ as downsampling and upsampling operators, a symmetric encoder-decoder architecture called U-Net \cite{Ronneberger2015} had proposed for biomedical and medical image segmentations. The U-Net works well on small data set due to
its mild parameter size and it can partly prevent the network from overfitting. Another encoder-decoder structure for image segmentation
is DeepLab series works, the introduced atrous convolution, atrous spatial pyramid pooling blocks can extract more multi-scale feature. In the earlier DeepLab, the CRF is as post-processing technique to smooth object boundaries, it has been shown that this spatial regularization
can improve the segmentation results.
However, this post-processing could not back propagate the regularization prior  to the parameter updating when training, and
it cannot correct some errors of network output.
To impose a spatial regularization, TV can be introduced in the network. Our previous work \cite{Jia2019}
replace the softmax activation function with a minimizer of TV regularized variational problem. This regularized softmax ensure the DCNN can produce piecewise constants outputs, which is suitable to image segmentation. However, TV is not smooth and it would case
gradient explosion in back propagation. To solve this problem, the dual algorithm \cite{Chambolle2005Total} of TV is be applied in \cite{Jia2019}. But it  needs hundreds of iterations to get the converged dual variable which contains spatial regularization information, this means that it needs hundreds of layers in DCNN and it would cost many computational resources.

\subsection{Variational segmentation methods}

\subsubsection{Potts model}
A well-known variational  image segmentation model is the Potts model. It solves:
\begin{equation}
\min_{\Omega_i}\sum_{i=1}^I \int_{\Omega_i} o_{i}(x) \dx+\lambda\sum_{i=1}^I |\partial\Omega_i|
\end{equation}
where$\cup_{i=1}^I\Omega_i=\Omega, \Omega_i\cap\Omega_{\hat{i}}=\varnothing$ is a segmentation condition, $I$ is the total classes, and $o_{i}(x)$ is a
similarity (feature) of pixel at $x$ in $i$-th class. The second term is a spatial regularization term and $\lambda>0$ is a parameter. Usually, $\Omega_i$ can be represented as its relax indicative function $u_i(x)\in[0,1]$ and the segmentation condition can be written as a simplex
\begin{equation}\label{simplexu}
\mathbb{U}=\{\bm{u}=(u_1,\cdots,u_I)\in[0,1]^I:\sum_{i=1}^I u_{i}(x)=1,\forall x\in\Omega\}.
\end{equation}
Many methods had developed based on this model. For example, let $\bm u$ be a level set representation, it would be the level set segmentation method\cite{chan2001active}. When $\bm u$ is binary, it is the graph-cut/max-flow based segmentation energy \cite{Rother2004,Yuan2010}. 
In many segmentation models \cite{chan2001active,Bresson2007,Yuan2010}, TV
was applied to represent the length term $|\partial\Omega_i|$.

\subsubsection{Threshold dynamics method for image segmentation}
Instead of TV regularization, a smooth regularization term derived from threshold dynamics was proposed in \cite{merriman1992diffusion,MBO2,EsedoThreshold} to approximate by boundaries length $|\partial\Omega_i|$:
\begin{equation}\label{eq:TD}
|\partial\Omega_i|\approx\sqrt{\frac{\pi}{\sigma}}\sum_{\hat{i}=1,\hat{i}\neq i}^{I}\int_{\Omega}u_i(x) (k*u_{\hat{i}})(x) \dx
\end{equation}
where $k$ is a Gaussian kernel
$k(x)=\frac{1}{2\pi\sigma^2}e^{-\frac{|x|^2}{2\sigma^2}}$
and the symbol $*$ is the convolution operator.
This regularization term penalizes the pixels that are isolated and imposed the spatial dependency into the model. It has been shown \cite{Miranda2007} that this threshold dynamics regularization $\Gamma$-converge to $|\partial\Omega_i|$ when $\sigma\rightarrow 0$.\\

Using the segmentation condition \eqref{simplexu}, the Potts model can be approximated by
\begin{equation}\label{th}
\min_{\bm{u}\in \mathbb{U}}\left\{\underbrace{\langle\bm o, \bm u\rangle}_{\mathcal{F}(\bm u;\bm o)}+\lambda\underbrace{\langle\bm u, k*(1-\bm u)\rangle}_{\mathcal{R}(\bm u)}\right\},
\end{equation}
where $\langle\bm o, \bm u\rangle=\sum_{i=1}^I \int_{\Omega} o_i(x)u_i(x)\dx$ and $\langle\bm u, k*(1-\bm u)\rangle=\sum_{i=1}^I \int_{\Omega} u_i(x)(k*(1-u_i))(x)\dx$
The difficulty in solving this minimization problem is that the regularizzation  term, which denotes as $\mathcal{R}(\bm u)$, is not linear. So, one solution is to linearize it as:
$$\hat{\mathcal{R}}(\bm u;\bm u^{t_1})= \langle\bm o, k*(1-\bm u^{t_1})\rangle,$$
where $t_1$ is the iteration number. This linearization method was adopted early and studied in \cite{Liu2011} from the perspective of constrained optimization. A similar idea has recently been studied in \cite{ICTM} as an iterative threshold method. With this linearization, the non-convex problem becomes a linear problem in each iteration. In real implementations, this algorithm is quite efficient and stable. It converges in several iterations
for most of the cases.

\section{The proposed method}\label{sec3}
\subsection{The motivation of the proposed method}\label{pm_motivation}
Let us analyze the classification function adopted in DCNN. Usually, a softmax layer $\left[\mathcal{S}(\bm o(x))\right]_i=\frac{\exp(o_i(x))}{\sum_{\hat{i}=1}^{I}\exp(o_{\hat{i}}(x))}$ is used to force the output of $\bm o$ to be a probability before cross entropy loss function. This step is very important for training since the popular cross entropy loss functional $-\sum_{x\in\Omega}\sum_{i=1}^I g_i(x)\ln o^{T}_i(x)$ in semantic segmentation can reach its minimization $\bm o^T=\bm g$ only when  the condition $\sum_{i=1}^Io^T_i(x)=1$ is satisfied. Here $\bm g(x)=(g_1(x),g_2(x),\cdots,g_i(x))$ is the indicative function (one-hot vector) of the ground truth. However, in the test step of DCNNs, the softmax layer may not be needed since we can use the
maximum operator
\begin{equation}\label{binaryu}
\widetilde{u}_i(x)=\begin{cases}
1, &i=\arg\max\{o_1(x),o_2(x),\cdots,o_I(x)\},\\
0, & else.
\end{cases}
\end{equation}
to get a binary prediction 
$$\widetilde{\bm u}(x)=(\widetilde{u}_1(x),\widetilde{u}_2(x),\cdots,\widetilde{u}_I(x)).$$

Compared to  variational models, the predictions of both the softmax and binary segmentation $\widetilde{\bm u}$ are
independent (pixel by pixel) with respect to spatial variable $x$ and without any spatial priors. However, many spatial priors such as
piecewise constants regions and shapes are important in semantic segmentation. To the best of our knowledge,
except for our previous work on spatial regularization \cite{Jia2019} and volume preserving \cite{Li2019},
there is no other work using DCNNs that can handle 
these kinds of shape prior. One of the difficulties comes from fact that the variational segmentation models and DCNNs are separated and many  existing techniques  such as volume preserving, shape prior, spatial regularization in variational models can not be  extended to DCNNs. Another difficulty is that the cost functionals in variational image segmentation models are often non-smooth.
If we use our techniques to include these non-smooth functionals into DCNNs,  the non-smooth variational problems would lead to gradient explosion risk during the back propagation process in DCNNs.

In the next, we shall provide a variational viewpoint for softmax activation function. With this framework, many existing techniques
in variational image segmentation models can be adopted into DCNNs. In our method, the softmax activation function can be regarded as a dual function in the primal-dual image segmentation model.

\subsection{Variational explanation for softmax}
In this section, we will give the softmax a variational interpretation. This enables us to  incorporate the techniques in many well-known variational methods into DCNNs.
It is easy to check that \eqref{binaryu} is a maximizer (may not be unique) of k-means type energy
\begin{equation*}
\widetilde{\bm u}=\underset{\bm{u}\in \mathbb{U}}{\arg\max}\left\{\mathcal{F}(\bm u;\bm o)=\langle\bm o, \bm u\rangle\right\},
\end{equation*}
and $\mathcal{F}(\widetilde{\bm u})=\int_{\Omega}\max \left\{o_1(x),o_2(x),\cdots,o_I(x)\right\} \dx.$

As mentioned earlier, $\widetilde{\bm u}$ is binary and its back propagation may not  be stable.
In order to get a smooth segmentation $\bm u$ for back propagation, we can smooth the $\max$ function as log-sum-exp function. For notional simplicity, we will use the following definition. 
\begin{definition}[log-sum-exp function]
\begin{equation*}\label{f_p}
\mathcal{M}_{\varepsilon} (\bm o)=\varepsilon ln \sum_{i=1}^I e^{\frac{o_i}{\varepsilon}}. 
\end{equation*}
\end{definition}
It is not difficult to verify $\underset{\varepsilon\rightarrow 0}{\lim}\mathcal{M}_{\varepsilon}(\bm o)=\max\{o_1,\cdots,o_I\}$ and $\mathcal{M}_{\varepsilon}$ is smooth and convex. One may notice that the derivative of $\mathcal{M}_{\varepsilon}$ with respect with to $o_i$ is exactly the softmax operation $\mathcal{S}$ when $\varepsilon=1$. This inspires us to find its dual representation. By standard convex analysis, we can get the following useful proposition:
\begin{pro}\label{pro1}
The Fenchel-Legendre transformation of $\mathcal{M}_{\varepsilon}$ is
\begin{equation*}\label{df_p}
\begin{array}{lll}
\mathcal{M}^{*}_{\varepsilon}(\bm u):=&\max\limits_{\bm o}\left\{\langle\bm o,\bm u\rangle-\mathcal{M}_{\varepsilon}(\bm o)\right\}\\
&=\left\{
\begin{array}{lll}
\varepsilon\langle\bm u,\ln \bm u\rangle,& \bm u\in\mathbb{U},\\
+\infty,& else.
\end{array}
\right.
\end{array}
\end{equation*}
where $\mathbb{U}$ is the same as in \eqref{simplexu}.
\end{pro}
\begin{pro}\label{pro2}
The twice Fenchel-Legendre transformation of $\mathcal{M}_{\varepsilon}$ is
\begin{equation*}
\mathcal{M}_{\varepsilon}^{**}(\bm o)=\max\limits_{\bm u\in\mathbb{U}}\left\{\langle\bm o,\bm u\rangle-\varepsilon\langle\bm u,\ln \bm u\rangle\right\}.
\end{equation*}
\end{pro}
Since $\mathcal{M}_{\varepsilon}(\bm o)$ is convex, and thus we have $\mathcal{M}_{\varepsilon}(\bm o)=\mathcal{M}_{\varepsilon}^{**}(\bm o)$.
Given the features $\bm o$ extracted by DCNN, instead of getting the binary segmentation cost functional $\mathcal{F}(\widetilde{\bm u})$ according to formulation \eqref{binaryu}, one can get a smooth version of $\mathcal{F}(\widetilde{\bm u})$. We write it as
 $\mathcal{F}_{\varepsilon}(\widetilde{\bm u})=\int_{\Omega} \mathcal{M}_{\varepsilon} \left\{o_1(x),o_2(x),\cdots,o_I(x)\right\} \dx.$ According to
 propositions \ref{pro1} and \ref{pro2}, we can get
 \begin{equation*}
 \begin{array}{rl}
 \widetilde{\bm u}=&\arg\max\limits_{\bm u\in\mathbb{U}}\left\{\langle\bm o,\bm u\rangle-\varepsilon\langle\bm u,\ln \bm u\rangle\right\}\\
 =&\arg\min\limits_{\bm u\in\mathbb{U}}\left\{\langle-\bm o,\bm u\rangle+\varepsilon\langle\bm u,\ln \bm u\rangle\right\}.
 \end{array}
 \end{equation*}
 The last equation follows by the fact that the
 maximizer and minimizer of
 these two problem are the same.
One can check that $\widetilde{\bm u}$ is the classical softmax operator $\mathcal{S}$ when $\varepsilon=1$.

From this variational viewpoint, softmax layer is a minimizer of the above functional. Compared with Potts model and its variants, there is an entropy term and this  enables the segmentation to be a soft threshold. This  is beneficial to back propagation in DCNN since it can ensure the
solution $\widetilde{\bm u}$ to be smooth.
On the other hand, the segmentation energy lacks spatial prior and it could be improved by adding spatial regularization as in many well-known variational image segmentation models. In our previous work \cite{Jia2019}, TV was integrated into softmax in this framework. 
In \cite{Jia2019}, the dual method is applied to overcome this difficult. However, tests show that it may need hundreds of iterations to calculate the dual variable in each training step. 
In variational methods, it has been experimentally proven (e.g. \cite{Liu2011,ICTM}) that the threshold dynamics is faster and more stable than many fast algorithm used for TV regularization. 
In the following, we will use 
threshold dynamics method for DCNNs.

Let us mention that the variational interpretation for softmax is an extension of the well-known Gaussian Mixture Model segmentation with  Expectation Maximum (EM) algorithm.
In fact, EM algorithm for GMM is a special case of the variational softmax problem when entropic parameter $\varepsilon=1$. To be more precise, $\mathcal{M}_{\varepsilon}$ is a log-likelihood function of  exponential mixture distribution model from statistical viewpoint and its optimization problem can be efficiently solved by the well-known EM algorithm. In this way, this interpretation for softmax can have another statistical theory.
We do not plan to show the details of this theory here.
Interested readers can find the intrinsic connections in our previous work \cite{liu2013weighted}.

\subsection{Proposed soft threshold dynamics method}
From the previous  analysis, one can see that the entropy regularization is good for
the back propagation of DCNNs since it can make the segmentation layer to be smooth.
To keep the spatial smooth prior, the threshold dynamics regularization which is smooth can be used.

\subsubsection{The proposed Model}
We propose the following soft threshold dynamics (STD) model:
\begin{equation*}
 \widetilde{\bm u}=\arg\min\limits_{\bm u\in\mathbb{U}}\left\{\underbrace{\langle-\bm o,\bm u\rangle+\varepsilon\langle\bm u,\ln \bm u\rangle}_{:=\mathcal{F}(\bm u;\bm o)}\right.
 \left.+\underbrace{\lambda\langle e\bm u, k*(1-\bm u)\rangle}_{:=\mathcal{R}(\bm u)}\right\}.
 \end{equation*}
Here $e\geqslant 0$ is a given weighting function such as image edge detection function $e(x)=\frac{1}{1+||\nabla \bm v^0(x)||}$. It can be shown that $\mathcal{R}(\bm u)\propto\sum_{i=1}^I\int_{\partial \Omega_i} e\mathrm{d}s$ when the kernel $k$ satisfies some mild conditions \cite{Liu2011}. If $e\equiv 1$, then $\mathcal{R}(\bm u)$
penalizes the length of boundaries. When $e$ is an image edge detection function, then $\mathcal{R}(\bm u)$ would be an active contour term which regularize the length of the contours.

\begin{remark}
When $\varepsilon=0, e \equiv 1$, our model would reduce to the threshold dynamics \eqref{eq:TD}.
However, the solution of the original threshold dynamics is binary and it would make the cross entropy loss to be infinity. What is more, its back propagation has a potential gradient explosion risk.
\end{remark}
In the next, we will develop an energy decay algorithm to fast minimize this problem.

\subsubsection{The proposed algorithm}
Since $\mathcal{R}$ is concave if the kernel function $k$ (e.g. Gaussian kernel) is semi-positive definite,
we can replace the concave functional $\mathcal{R}$ with its supporting hyperplane in the minimization problem and
get an iteration
\begin{equation}\label{STDiteration}
\begin{array}{rl}
{\bm{u}}^{t_1+1}&=\underset{\bm{u}\in\mathbb{U}}{\arg\min}\left\{\mathcal{F}(\bm u;\bm o)+\mathcal{R}(\bm{u}^{t_1})+\langle \bm p^{t_1},\bm{u}-\bm{u}^{t_1} \rangle\right\}.\\
\end{array}
\end{equation}
Here $\bm p^{t_1}=\lambda((k*(1-\bm u^{t_1}))e-k*(e\bm u^{t_1}))\in\partial \mathcal{R}({\bm{u}}^{t_1})$ and $\partial \mathcal{R}({\bm{u}}^{t_1})$ is the subgradient of the concave functional $\mathcal{R}$ at $\bm u^{t_1}$. More details of calculating $\bm p^{t_1}$ can be found in appendices \ref{appendixA}.
When the weighting function $e=1$, then
$\bm p^{t_1}=\lambda k*(1-2\bm u^{t_1})$.
For this iteration \eqref{STDiteration}, we have the following unconditional stability for the energy:
\begin{thm}\label{theo1}
(Energy descent). Let $\bm{u}^{t_1}$ be the $t_1$-th
iteration of \eqref{STDiteration}, then we have
$\mathcal{F}({\bm{u}}^{t_1+1})+\mathcal{R}({\bm{u}}^{t_1+1})\leq \mathcal{F}({\bm{u}}^{t_1})+\mathcal{R}({\bm{u}}^{t_1}).$
\end{thm}
\begin{IEEEproof}
Please see the details in the appendices \ref{appendixB}.
\end{IEEEproof}

Moreover, the problem \eqref{STDiteration} has a softmax solution
 $$u_i^{t_1+1}(x)=\frac{e^{\frac{o_i(x)-\lambda\left((k*(1-u_i^{t_1}))e-k*(eu_i^{t_1})\right)(x)}{\varepsilon}}}{\sum^I_{\hat{i}=1} e^{\frac{o_{\hat{i}}(x)-\lambda\left((k*(1-u_{\hat{i}}^{t_1}))e-k*(eu_{\hat{i}}^{t_1})\right)(x)}{\varepsilon}}}.
$$

Compared to the classical softmax, the main difference is that the solution given by the above formula  has a spatial prior represented by term $\lambda \left((k*(1-\bm u))e-k*(e\bm u)\right)$, which implies
the smooth spatial prior can be easily plugged  into DCNN architectures by adding several convolutions between the softmax output in the previous layers and kernel $k$. For classical variational models, this scheme is very stable and converge fast in real implementation. This algorithm extends the gradient decent algorithm for TV regularization of Algorithm 1 of \cite{Bae2011} to TD with soft thresholding.  In the overwhelming majority cases, it would be converged within 10 iterations. This is the main advantage of STD compared to TV regularization.

We summary the STD softmax segmentation in the algorithm \ref{alg1}.

\begin{algorithm}
\caption{STD-softmax }\label{alg1}
\KwIn{The feature $\bm o$}
\KwOut{Soft segmentation function $\bm u$.}
\textbf{Initialization:} $\mathbf{u}^0=\mathcal{S}(\bm o).$\\
\For{$t_1=0,1,2,\cdots$}{
1. compute the solution of \eqref{STDiteration} by STD-softmax
$$
 \bm u^{t_1+1}=\mathcal{S}\left(\frac{\bm o-\bm p^{t_1}}{\varepsilon}\right).
$$\\
2. Convergence check. If it is converged, end the algorithm.\\
}
\Return Segmentation function $\boldsymbol{u}$.
\end{algorithm}

\subsection{Volume preserving soft threshold dynamics}
In the previous discussion, the set $\mathbb{U}$ is a simplex, which stands for an image segmentation condition and it lacks volume prior. We can easily
add  a volume constraint by modifying $\mathbb{U}$ as
$$\mathbb{U}_{\bm V}=\left\{\bm u\in \mathbb{U}:~~\int_{\Omega}u_i(x)\dx=V_i\right\},$$
where $\bm V=(V_1,V_2,\cdots,V_I)$ is a given volume in which $V_i$ stands for the volume of $i$-th class.
Thus the Volume Preserving Soft Threshold Dynamics (VP-STD) model can be written as
\begin{equation*}
 \widetilde{\bm u}=\arg\min\limits_{\bm u\in\mathbb{U}_{\bm V}}\left\{\mathcal{F}(\bm u;\bm o)+\mathcal{R}(\bm u)\right\}.
 \end{equation*}
Similarly as \eqref{STDiteration}, its related iteration could be
\begin{equation}\label{VPSTDiteration}
{\bm{u}}^{t_1+1}=\underset{\bm{u}\in\mathbb{U}_{\bm V}}{\arg\min}\left\{\mathcal{F}(\bm u;\bm o)+\underbrace{\langle \bm{u},\bm p^{t_1}\rangle}_{:=\hat{\mathcal{R}}(\bm u)}\right\}.
\end{equation}

Due to the  volume constraint in $\mathbb{U}_{\bm V}$, there is no closed-form minimizer for \eqref{VPSTDiteration}. However,
it can be regarded as an entropic regularized optimal transport \cite{cuturi2016smoothed} and can be solved efficiently by dual algorithm using the following fact:
\begin{pro}\label{pro3}
The primal problem \eqref{VPSTDiteration} has an equivalent dual problem
\begin{equation}
\underset{\bm u\in\mathbb{U}_{\bm V}}{\min}\left\{\mathcal{F}(\bm u;\bm o)+\hat{\mathcal{R}}(\bm u)\right\}=\mathop{\mathrm{max}}_{\bm q}\left\{\langle\bm q,\bm V\rangle+\langle\bm q^{c,\varepsilon},\bm 1\rangle-|\Omega|\varepsilon\right\},
\end{equation}
where $\bm q^{c,\varepsilon}=(q^{c,\varepsilon}_1,q^{c,\varepsilon}_2,\cdots,q^{c,\varepsilon}_I)$ is a $\varepsilon$ entropic regularization c-concave transform of $\bm q$ defined by
$$\bm q^{c,\varepsilon}(x)=-\mathcal{M}_{\varepsilon}(\bm o(x)-\bm p^{t_1}(x)+\bm q(x)).$$
In addition, the optimizers of primal and dual problem $\widetilde{\bm u}$ and $\widetilde{\bm q}$ have the relation $\widetilde{\bm u}=\mathcal{S}\left(\frac{\bm o-\bm p^{t_1}+\widetilde{\bm q} }{\varepsilon}\right)$.
\end{pro}
\begin{IEEEproof}
The detailed proof can be found in appendices \ref{appendixC}.
\end{IEEEproof}

According to the proposition \ref{pro3}, we can get $\bm u^{t_1+1}$ by solving dual problem about $\bm q$.
The related iteration could be
\begin{equation} \label{vpstd_qupdate}
\bm q^{t_2+1}=
\bm q^{t_2}+\varepsilon 
\left(\log V_i-\log\displaystyle\biggl.\int_{\Omega}\frac{e^{\frac{o_i(x)-p_i^{t_1}+q_i^{t_2}(x)}{\varepsilon}}}{\sum\limits^I_{\hat{i}=1}e^{\frac{o_{\hat{i}}(x)-p_{\hat{i}}^{t_1}+q_{\hat{i}}^{t_2}(x)}{\varepsilon}}}\dx\right),
\end{equation}
where $t_2$ is an inner iteration number.

This idea is related to the logarithmic Sinkhorn iteration\cite{Cuturi2013Sinkhorn}, and the convergence has been proven in \cite{Franklin1989On}. When we get a converged $\widetilde{\bm q}$,
then $\bm u^{t_1+1}$ in problem \eqref{VPSTDiteration} can be obtained according to proposition \ref{pro3}.
In real implementations, 
we just let $t_2=1$ and only do $1$ inner iteration for this volume constraint and we empirically find that it can get the desirable segmentation results.
We summarize the VP-STD softmax algorithm in algorithm \ref{alg2}:
\begin{algorithm}
\caption{VP-STD softmax }\label{alg2}
\KwIn{The feature $\bm o$ and volume $\bm V$.}
\KwOut{Volume preserving soft segmentation function $\bm u$.}
\textbf{Initialization:} $\mathbf{u}^0=\mathcal{S}(\bm o), \bm q^0=0.$\\
\For{$t=0,1,2,\cdots$}{
1. calculate the volume preserving dual variable by \eqref{vpstd_qupdate}, i.e.
\begin{equation*}
\bm q^{t_1+1}=\bm q^{t_1}+\varepsilon\left(\log \bm V
-\log \langle \mathcal{S}(\frac{\bm o-\bm p^{t_1}+\bm q^{t_1}}{\varepsilon}),\bm 1\rangle \right).
\end{equation*}

2. compute the solution of \eqref{VPSTDiteration} by VP-STD softmax
$$
 \bm u^{t_1+1}=\mathcal{S}\left(\frac{\bm o-\bm p^{t_1}+\bm q^{t_1+1}}{\varepsilon}\right).
$$\\
3. Convergence check. If it is converged, end the algorithm.\\
}
\Return Segmentation function $\boldsymbol{u}$.
\end{algorithm}

\begin{remark}
for large $\varepsilon$, $u_i$ would not be the indicative function of $\Omega_i$ and $\int_{\Omega} u_i(x)\dx$ would not be the exact volume. However, it is enough to improve the segmentation accuracy in many applications if there is a volume priori.
\end{remark}

\subsection{Star-shape prior soft threshold dynamics}
\begin{figure}
\includegraphics[width=1.0\linewidth]{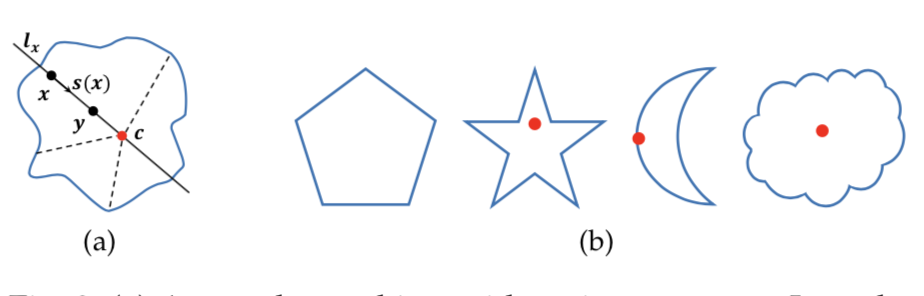}
\caption{(a) A star-shape object with a given center $c$. Let $x$ be inside the object, then for any $y$ on the directed straight line from $x$ to $c$ is also inside the object. (b) Star-shape examples. A convex shape (the above regular pentagon) with an inner center must be a star-shape.}\label{fig:1}
\end{figure}

Star-shape prior has been  widely used in many variational image segmentation models, especially in the field of medical image segmentation. Many of the medical segmentation results are required to be star-shape.
A star-shape object has the following proposition: if $x$ is inside the object, then all the points lie on the direct line from $x$ to a given center point $c$ are also inside the object. The center point $c$ can be called the center of the star-shape. In \figurename \ref{fig:1}, we have supplied some examples of star-shapes and their centers. 

If  $u:\Omega\rightarrow\{0,1\}$ is an indicative function of an object,
a discrete energy of graph cut had been given in \cite{Ss} to force $u$ as an indicative function of a star-shape:
\[
\mathcal{P}_0(u(x),u(y))=\begin{cases}
0, &if~u(y)=u(x),\\
+\infty, &if~u(y)=0 ~and~ u(x)=1,\\
\beta\geqslant 0, &if~u(y)=1 ~and~ u(x)=0.
\end{cases}
\]
When minimizing $\mathcal{P}_0$, the neighborhood structure of $u$ with  $1-0-1$ would be avoided and thus force the object to be a star-shape.
To integrate the star-shape into DCNNs, we need a continuous condition for soft threshold segmentation. In the next, we will
derive a continuous star-shape energy constraint for softmax.

Let $\alpha=+\infty$, for binary segmentation, we have
\[
\mathcal{P}_0(u(x),u(y))=\max \{\alpha(u(x)-u(y)),\beta(u(y)-u(x))\}
\]
where $u(x),u(y)\in \{0,1\}$.\\

Denote $c$ to be a given center of a star-shape object, for all $x$ in the object, we denoted the unit directed vector from $x$ to
$c$ as $\bm s(x)$. Besides, we let $l_{x}$ be a directed straight line starting at $x$ with the direction $\bm s(x)$.
Let us recall that the star-shape energy can be written as
\[
\mathcal{P}(u)=\sum_{x\in\Omega}\sum_{y\in{l_x}}\mathcal{P}_0(u(y),u(x)).
\]
By the definition of directional derivative
\[
u(y)-u(x)\approx\frac{\partial u}{\partial \bm {s}} (y-x)=\langle\nabla u(x),\bm{s}(x)\rangle(y-x)
\]

For simplification, we can let $y-x=1$, which implies that we always choose $y$ at the line $l_x$ located one unit far away from $x$.
Then we have
\[
u(y)-u(x)\approx\langle\nabla u(x),\bm{s}(x)\rangle.
\]

Substituting the above equation into $\mathcal{P}_0$, we have
\begin{equation*}
\mathcal{P}(u)=\int_{\Omega} \max\left\{-\alpha\langle\nabla u(x),\bm{s}(x)\rangle,\beta \langle\nabla u(x),\bm{s}(x)\rangle\right\}.\\
\end{equation*}
Let us simplify it again, if $\beta=0$, then
\begin{equation}
\mathcal{P}(u)=\begin{cases}
0, & \langle\nabla u(x),\bm{s}(x)\rangle\geq 0,\\
+\infty, &\langle \nabla u(x),\bm{s}(x)\rangle < 0.
\end{cases}
\end{equation}
This is an indicative function of a convex set
$$\mathbb{P}=\{u:~\langle \nabla u(x),\bm{s}(x)\rangle\geq 0\}.$$

Therefore, the proposed star-shape soft threshold dynamics can be
\begin{equation*}
 \widetilde{\bm u}=\arg\min\limits_{\bm u\in\mathbb{U}}\left\{\mathcal{F}(\bm u;\bm o)+\mathcal{R}(\bm u)+\mathcal{P}(u_i)\right\},
\end{equation*}
in which $u_i$ is the segmentation function of $i$-th region which has to be star-shape.

Similarly, we have the dual problem in terms of KKT condition
\begin{equation*}
 (\widetilde{\bm u},\widetilde{q})=\arg\min\limits_{\bm u\in\mathbb{U}}\max\limits_{q\geqslant 0}\left\{\mathcal{F}(\bm u;\bm o)+\mathcal{R}(\bm u)-\langle q,\bm s\cdot \nabla u_i\rangle\right\}.
\end{equation*}

By linearizing $\mathcal{R}(\bm u)$ as $\hat{\mathcal{R}}(\bm u)$, the above saddle problem can be solved by the following alternating minimization scheme:

\begin{equation*}
\left\{
\begin{array}{rl}
 q^{t_1+1}=&\max\{q^{t}-\tau_q\bm s\cdot \nabla u_i^{t},0\}.\\
\bm u^{t_1+1}=&\underset{\bm{u}\in\mathbb{U}}{\arg\min}\left\{\mathcal{F}(\bm u;\bm o)+\hat{\mathcal{R}}(\bm u)+\langle div(q^{t_1+1}\bm s), u_i\rangle\right\},\\
\end{array}
\right.
\end{equation*}
Here $div$ is the divergence operator and $\tau_q$ is a small step. Larger $\tau_q$ can make
the object to be star-shape quickly, but it has the risk 
that the algorithm is not stable. We empirically find that
$\tau_q$ can be set large when $\varepsilon$ is large. Thus,
we choose $\tau_q=\varepsilon$ in this paper.
The $\bm u$-subproblem has a closed-form softmax solution which can be found in the Star-Shape Soft Threshold Dynamics (SS-STD) softmax segmentation algorithm \ref{alg3}.


\begin{algorithm}
\caption{SS-STD softmax}\label{alg3}
\KwIn{The feature $\bm o$, and a center $c$ of star-shape.}
\KwOut{Soft segmentation function $\bm u$.}
\textbf{Initialization:} $\mathbf{u}^0=\mathcal{S}(\bm o)$. Calculating the star-shape vector field $\bm s$ according to $c$.\\
\For{$t_1=0,1,2,\cdots$}{
1.update dual variable for the $i$-th star-shape region
$$q^{t_1+1}=\max\{q^{t_1}-\tau_q\bm s\cdot \nabla u_i^{t_1},0\}.$$
2. compute the solution of \eqref{STDiteration} by SS-STD softmax
$$
 \bm u^{t_1+1}_{\hat{i}}=\mathcal{S}\left(\frac{\bm o_{\hat{i}}-\bm p^{t_1}_{\hat{i}}-\delta_{\hat{i},i}div(q^{t_1+1}\bm s)}{\varepsilon}\right),~\hat{i}=1,\cdots,I.
$$\\
3. Convergence check. If it is converged, end the algorithm.\\
}
\Return Segmentation function $\boldsymbol{u}$.
\end{algorithm}

\begin{remark}
The proposed SS-STD softmax can be regarded as a soft TD extension for the star-shape methods of \cite{Ss,Yuan2012}.
Let $\varepsilon\rightarrow 0$ and the regularization term $\mathcal{R}$ be an  anisotropic discrete TV, the SS-STD softmax would be equivalent to the discrete min-cut star-shape method \cite{Ss}. With the dual representation and isotropic TV, it would be reduced to the continuous max-flow star-shape method \cite{Yuan2012}.
Both of these two star-shape  methods belong to the binary segmentation and they cannot be directly plugged into the DCNN architecture due to the non-smoothness of the segmentation.
\end{remark}
\subsection{The STD softmax  block
for DCNN}
\begin{figure*}
\includegraphics[width=1.0\linewidth]{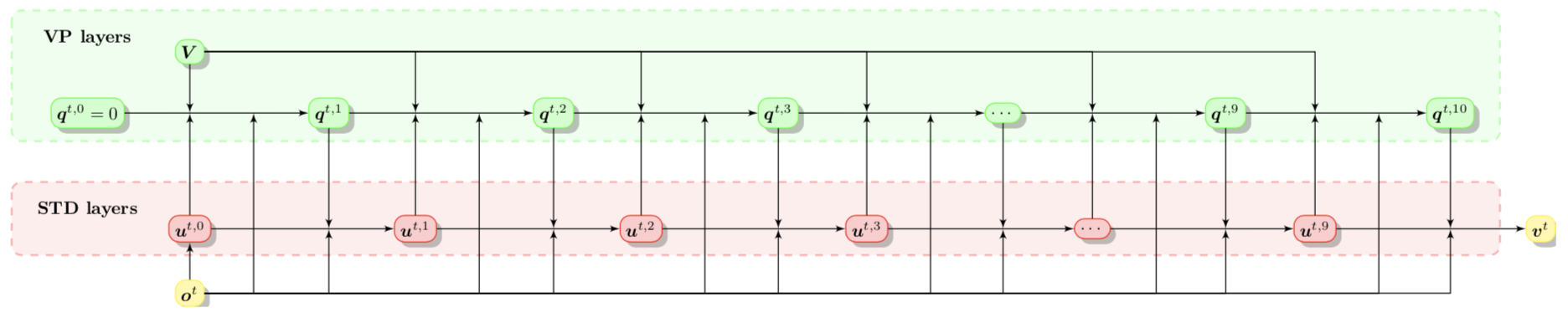}
\caption{The network architecture of the proposed STD and VP-STD softmax block. The red dash rectangle denotes the dual variables of regularization space; The green dash rectangle denotes the dual variables of volume preserving space.}\label{fig2}
\end{figure*}

Since the similarity term $\mathcal{F}$ and regularization term are both smooth, the calculations with them  in connections with the back propagations would be stable if we unroll the iteration algorithm as DCNN layers.
We can get a DCNN with a new soft thresholding dynamics (STD) for image segmentation as
 \begin{equation}\label{DCNN-STD}
\left\{
\begin{array}{rl}
\bm o^t=&\mathcal{T}_{\bm \Theta^{t-1}}(\bm v^{t-1},\bm v^{t-2},\cdots,\bm v^0),\\
\bm v^{t}=&\underset{\bm u\in\mathbb{U}}{\arg\min}\left\{\mathcal{F}(\bm u;\bm o^t)+\mathcal{R}(\bm u)\right\}, t=1,\cdots, T.
\end{array}
\right.
\end{equation}
Compared to the given activation function in original simple DCNN \eqref{eq:nn1}, here the proposed activation function
is a minimizer of a variational problem. This enable us to
use spatial variational prior in DCNNs  for segmentation. To get $\bm v^t$, several sublayers related to the algorithm \ref{alg1} should be added in the network. In this paper,
we set the number of the sublayers as $t_1=10$.
 As visualization,
we show these sublayers in \figurename \ref{fig2}. In this figure, the connections of STD layers are displayed in the red dash rectangle. Their strict mathematical relationships
can be found in algorithm \ref{alg1}.
Compared to the original softmax layer, the STD block incorporates spatial regularization, and the information can be transmitted between the original softmax and its dual variables. Therefore, the output $v^t$ should have spatial regularity.

\subsection{The Volume Preserving STD (VP-STD)  softmax  block
for DCNN}
Similar to the pervious section, the VP-STD algorithm \ref{alg2} also can be unrolled as a VP-STD softmax
block. It also can be plugged into the DCNN as
 \begin{equation}\label{DCNN-VP-STD}
\left\{
\begin{array}{rl}
\bm o^t=&\mathcal{T}_{\bm \Theta^{t-1}}(\bm v^{t-1},\bm v^{t-2},\cdots,\bm v^0),\\
\bm v^{t}=&\underset{\bm u\in\mathbb{U}_{\bm V}}{\arg\min}\left\{\mathcal{F}(\bm u;\bm o^t)+\mathcal{R}(\bm u)\right\}, t=1,\cdots, T.
\end{array}
\right.
\end{equation}

We show the VP-STD softmax block in the \figurename\ref{fig2}. The related strict mathematical relationships of variables can be found in Algorithm \ref{alg2}.
From this figure, one can find that the dual variables of Volume Preserving (VP) are contained in a green dash rectangle. They can enable the output of the DCNNs have
the volume prior.
Combing with STD and VP layers, it can produce segmentation results with volume constraint and spatial regularization.

\subsection{The Star-Shape STD (SS-STD)  softmax  block
for DCNN}
As for the SS-STD sublayers, it is slightly different from the
previous VP-STD sublayers according to the Algorithm \ref{alg3}. We can get a DCNN with star-shape output as
\begin{equation}\label{DCNN-SS-STD}
\left\{
\begin{array}{rl}
\bm o^t=&\mathcal{T}_{\bm \Theta^{t-1}}(\bm v^{t-1},\bm v^{t-2},\cdots,\bm v^0),\\
\bm v^{t}=&\underset{\bm u\in\mathbb{U},u_i\in\mathbb{P}}{\arg\min}\left\{\mathcal{F}(\bm u;\bm o^t)+\mathcal{R}(\bm u)\right\}, t=1,\cdots, T.
\end{array}
\right.
\end{equation}

\begin{figure*}
\includegraphics[width=1.0\linewidth]{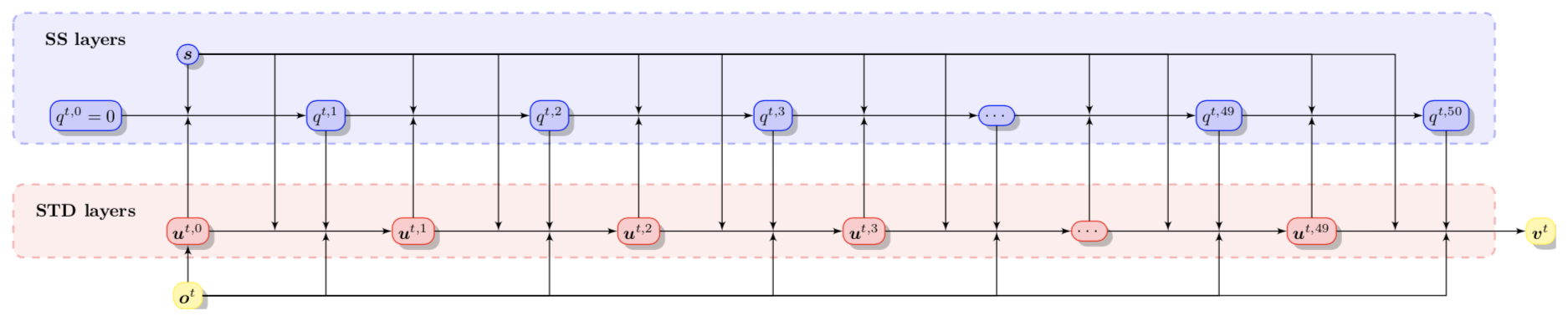}
\caption{The network architecture of the proposed SS-STD softmax block. The red dash rectangle denotes the dual variables of regularization space; The blue dash rectangle denotes the dual variables on a star-shape space.}\label{fig3}
\end{figure*}

We show the structure of SS-STD in \figurename\ref{fig3}.  In this figure, the blue dash rectangle contains the dual variables on the star-shape space, together with STD, they can produce results with star-shape and smooth boundaries. 
We test the algorithm \ref{alg3}, it would usually need 
at least hundreds of iterations to reach convergence.
To save computational source, we just take $t_1=50$ layers in the network to keep the star-shape prior. Experimental 
tests show that it still can improve segmentation accuracy even though in this case. Other fast algorithms can be developed to fast solve the star shape constraint, but we do not try this in our current implementation.

If preferred,  one can take the volume prior, star-shape prior and spatial regularization all together and get a DCNN block that can handle all these  spatial priors.

\subsection{Applying the proposed STD softmax based layers to DeeplabV3+}
To show how to apply the proposed method to the segmentation DCNNs and to see its performance, we take the popular DeepLabV3+
as the baseline. Mathematically, the choice of basic networks can be obtained by choosing different special operators $\mathcal{T}^{t}$ for each layers in \eqref{DCNN-STD} \eqref{DCNN-VP-STD}, and \eqref{DCNN-SS-STD}. Therefore, our method can be applied to any image segmentation networks such as U-Net \cite{Ronneberger2015} and many segmentation DCNN variants such as FastFCN\cite{FastFCN}, DA-RNNs\cite{xiang2017rnn}, ReSeg\cite{visin2016reseg} \emph{etc.}. 

\begin{figure*}
\includegraphics[width=1.0\linewidth]{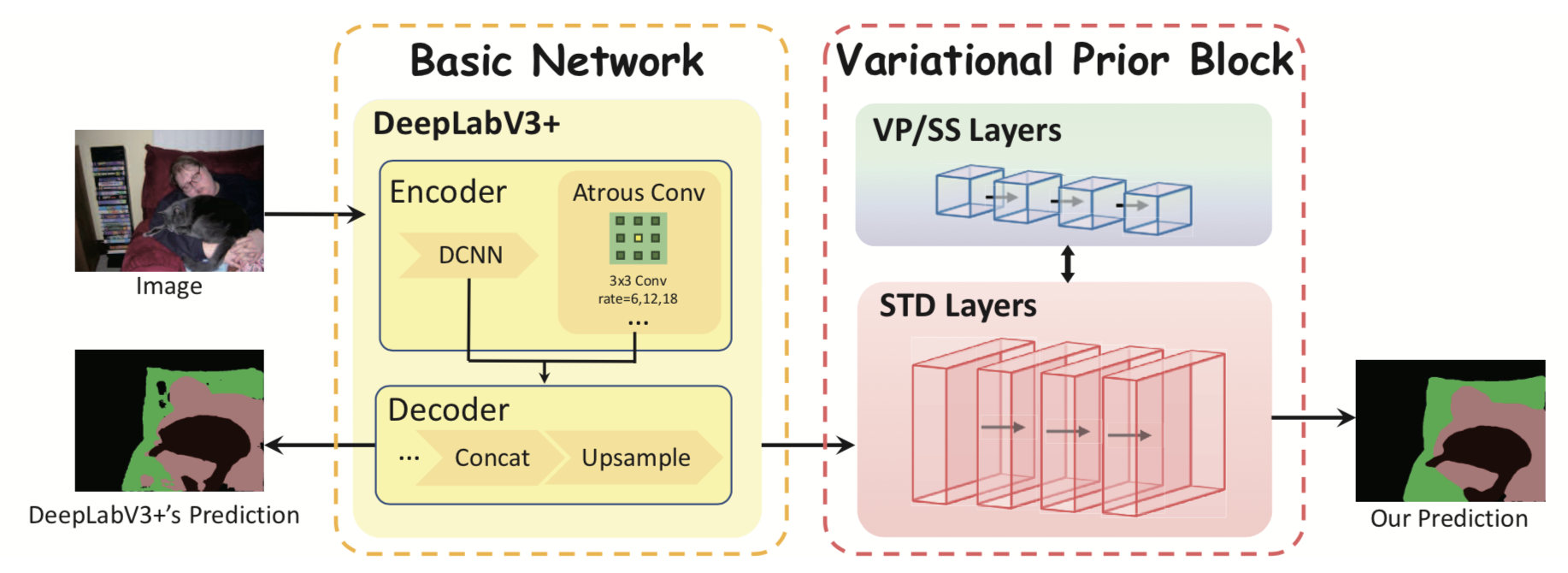}
\caption{The network architecture of STD -DeepLabV3+.
The outputs of this network  could have many spatial priors such as smooth boundaries, volume constraint, shape prior.}\label{fig4}
\end{figure*}

To simplify the computation, in this paper, we just replace the last softmax layer (softmax loss in the original DeepLabV3+ tensorflow implementation) with our proposed STD based softmax blocks. All other structures of the networks are remained as the same as DeepLabV3+'s. The loss function is chosen as the
softmax loss. The flowchart of the modified  DeepLabV3+ called STD-DeepLabV3+ is shown in \figurename\ref{fig4}. 

\section{Numerical experiments}\label{sec4}

\subsection{A toy experiment}
\begin{figure}
\includegraphics[width=1.0\linewidth]{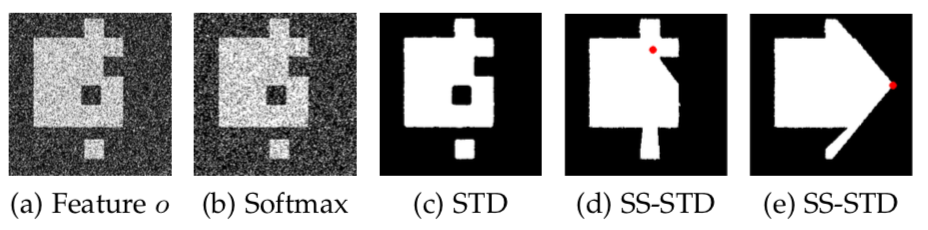}
\caption{Segmentation results by softmax, proposed STD-softmax (alg.\ref{alg1}) and SS-STD softmax (alg. \ref{alg3}). The red points in the last two figures are the given centers of the star-shape.}\label{fig5}
\end{figure}

To intuitively see the segmentation results by our method, a toy experiment to show the segmentation results by different algorithms can be found in \figurename~\ref{fig5}. In this figure, the first image $v$ 
is an object with noise. We segmented it into 2 classes in terms of different algorithms. The feature $\bm o$ is produced by $o_i=-\frac{1}{2}||v-\mu_i||^2$, where $\mu_i$
is the mean of $i$-th region which can be calculated by a K-means initialization. The last four images are the segmentation results produced by 3 different algorithms. Compared to 
softmax, the STD-softmax (alg. \ref{alg1}) can give a smooth boundary and is robust for noise. SS-STD softmax (alg. \ref{alg3}) can ensure the segmented regions to be star-shapes.

\subsection{Implementation details}
In this paper, we use tensorflow tool to implement our algorithms. As can be found in algorithms \ref{alg1} and \ref{alg2}, $(k*(1-\bm u^{t_1}))e-k*(e\bm u^{t_1})$ is 
a depth-wise convolution and it can be efficiently implemented by
``tf.nn.depthwise\_conv2d" function in tensorflow.
Other layers of STD, VP-STD and SS-STD are all simple operators which have been defined in tensorflow. Thus the
related backward propagation formulations are standard and can be automatically calculated by tensoflow.

In all the experiments, we set the weight function $e=1$ and
the spatial smoothness prior $(k*(1-\bm u^{t_1}))e-k*(e\bm u^{t_1})$ is reduced to a very simple representation $k*(1-2\bm u^{t_1})$. We believe that adding the image edges to the DCNN by choosing a suitable weighting $e(x)$
would be helpful for many applications. We do not test the effect of weight function $e$ in this paper.

We choose the kernel $k$ as a Gaussian kernel with support set size $7\times 7$ and standard deviation $\sigma=5.0$. Besides, $k$ is frozen and does not be updated (learned) in training. Let us mention that the size of $k$ may affect the smoothness of the segmentation results. Larger size would produce more smooth boundaries. $5-9$
is good enough  for most of the cases. 
Intuitively, learn of $k$ may slightly improve the results, but it will cost many computation resource. We do not test this in the experiment.

The codes of DeepLabV3+ is download from the official 
implementation \url{https://github.com/tensorflow/models/tree/master/research/deeplab}. For the baseline DeepLabV3+, we use this code to implement.
In all the following experiments, we choose Xception65 as 
the backbone network for DeepLabV3+ and STD-DeepLabV3+. For the DeepLabV3+, we use all the default setting in the codes. In the STD-DeepLabV3+, the proposed STD based blocks are plugged before softmax loss to keep the spatial priors. All the hyper-parameters
including learning rate are set as the same as DeepLabV3+'s.

As for the VP-STD method, the exact volume $\bm V$ can be calculated from the ground truth in the training. In this paper, we use the exact volume constraints in the prediction
to evaluate the efficiency of the algorithm. This may be not 
applicable in real applications. In real applications, one may guess an approximate volume ratio for prediction. Another 
alternative would be  to let the volume constraints to be box constraint with an estimate of the upper and lower bound for the volumes.  This is related to the unbalanced optimal transport theory. Similar algorithm can be designed according to the idea of this paper.

For star-shape prior, since the general public datasets such as 
PASCAL VOC 2012 and CITYSCAPE do not contain a class of objects that are fully star-shapes, we do not test our methods
on these datasets.
For many medial image processing datasets, star-shapes are  common. We have tested our algorithms on these datasets. 
We take a medical image dataset called ISIC2018 to show the performance of the SS-STD method. In both of the training and prediction stages, the center $c$ of the star-shape are all given in advance.

Our computation platform is a linux server equipped with $4\times 32$G Tesla V100-SXM2  GPUs.

\subsection{Algorithm evaluations}
In this section, we will evaluate the proposed method and compare with some of state-of-the-art methods on several datasets. 
\subsubsection{PASCAL VOC 2012}
This section is to evaluate the performance of the STD-softmax layers. The PASCAL VOC 2012 dataset \cite{everingham2010pascal} includes $1464$ training, $1449$ validation and $1456$ test images. In which it contains 20 foreground object classes and one background class. To accelerate the training and fair comparison, we use the pre-training weight ``xception65\_coco\_voc\_trainaug" from 
\url{https://github.com/tensorflow/models/blob/master/research/deeplab/g3doc/model_zoo.md} for both DeepLabV3+ and the modified STD-DeepLabV3+.
Similar to Deeplab, the crop size is chosen as $513\times 513$. We train the networks on training and validation sets (2913 images) and then apply them to the test set for prediction.

For fair comparison, we use the same parameters for DeeplabV3+ and the STD version: the batch size of training is set as
$16$, and the iteration number is $50000$. The learning rate is set as $0.0001$. 

The mean intersection-over-union (mIoU) across the 21 classes is employed to measure the performance of the algorithm. 

For this entropic parameter $\varepsilon$, it controls the sharpness of the binary functions of the segmentation, larger $\varepsilon$ can enable the algorithm to be stable but
produce fuzzy segmentations. For very small $\varepsilon$,
the segmentation $\bm u$ would be nearly binary and it may case the backward propagation to be unstable. Usually, this parameter can be set in $[0.1,3]$. We list the effects of this
parameter for training accuracy in the proposed STD-DeepLabV3+ in \tablename~\ref{tab1}. For this dataset, we numerically find that small $\varepsilon$ can improve the training accuracy.
\begin{table}[htp]
\caption{The mIoU of validation set in the training for different  entropic parameter $\varepsilon$ in the proposed STD-softmax. The  regularization parameter is fixed as $\lambda=1.0$.}\label{tab1}
\begin{center}
\begin{tabular}{cccccccc}
\toprule[2pt]
$\varepsilon$&&0.1&0.4&0.7&1.0&2.0&3.0\\
\hline
mIoU&&$\bm{94.40}$&$93.52$&$92.75$&$92.28$&92.03&91.79\\
\bottomrule[2pt]
\end{tabular}
\end{center}
\end{table}

 Another parameter in the proposed method is the regularization parameter $\lambda$, it balances the feature 
extracted by DCNN and spatial regularization term. Generally speaking, the larger $\lambda$, the more spatial regularization. $\lambda$ can be set in $[0,1.25]$. When $\lambda=0$, the proposed STD method would reduced to
the original softmax activation layer. The training accuracy affected by $\lambda$ are listed 
in \tablename~\ref{tab2}.
\begin{table}[htp]
\caption{The effects of regularization parameter $\lambda$ in the proposed STD-DeepLabV3+. The entropic parameter is fixed as $\varepsilon=0.1$.}\label{tab2}
\begin{center}
\begin{tabular}{cccccc}
\toprule[2pt]
$\lambda$&&0.5&0.75&1.0&1.25\\
\hline
mIoU&&$94.09$&$94.17$&$\bm{94.40}$&$92.90$\\
\bottomrule[2pt]
\end{tabular}
\end{center}
\end{table}

A possible method to automatically choose these parameters is to learn $\varepsilon$ and $\lambda$, but we do not pursue this possibility in the current paper. In the experiments, we just simply choose several  experiential parameters for test.

The \tablename~\ref{tab3} shows the mIoU of validation set 
for several state-of-the-art DCNNs method for segmentation.
Compared with the existing methods, the proposed
STD based DeepLabV3+ can achieve higher mIoU than others. The performance is improved from $91.34\%$ (DeepLabV3+) to $94.40\%$ and $95.21\%$.
It has about $2.9\%-3.7\%$ improvements.
We do not take the SS-STD method for comparison since not all the 21 classes objects are star-shapes.
We take our recent works TV-DeepLabV3+ \cite{Jia2019}
and VP-TV-DeepLabV3+ \cite{Li2019} for comparison, in which the TV regularization \cite{Jia2019} and volume preserving \cite{Li2019} were employed in these methods.
One can find the proposed STD based method can reach higher mIoU than TV's because the dual algorithm for TV
often needs hundreds of iteration to reach convergence. But only dozens of layers (30) were adopted in \cite{Jia2019,Li2019} due to the limitation of the computation resource and computational time.

To show the computation efficiency, we list the training speed for DeepLabV3+, TV, VP-TV, STD, VP-STD DeepLabV3+ in \tablename~\ref{tab5}. In this table, the iteration numbers per second for 5 methods are listed. The proposed STD-DeepLabV3+ is slower about $16\%$ than DeepLabV3+.
But it is faster about $30\%$ than TV-DeepLabV3+. The similar conclusion can be found in VP versions.

\begin{figure}
\includegraphics[width=1.0\linewidth]{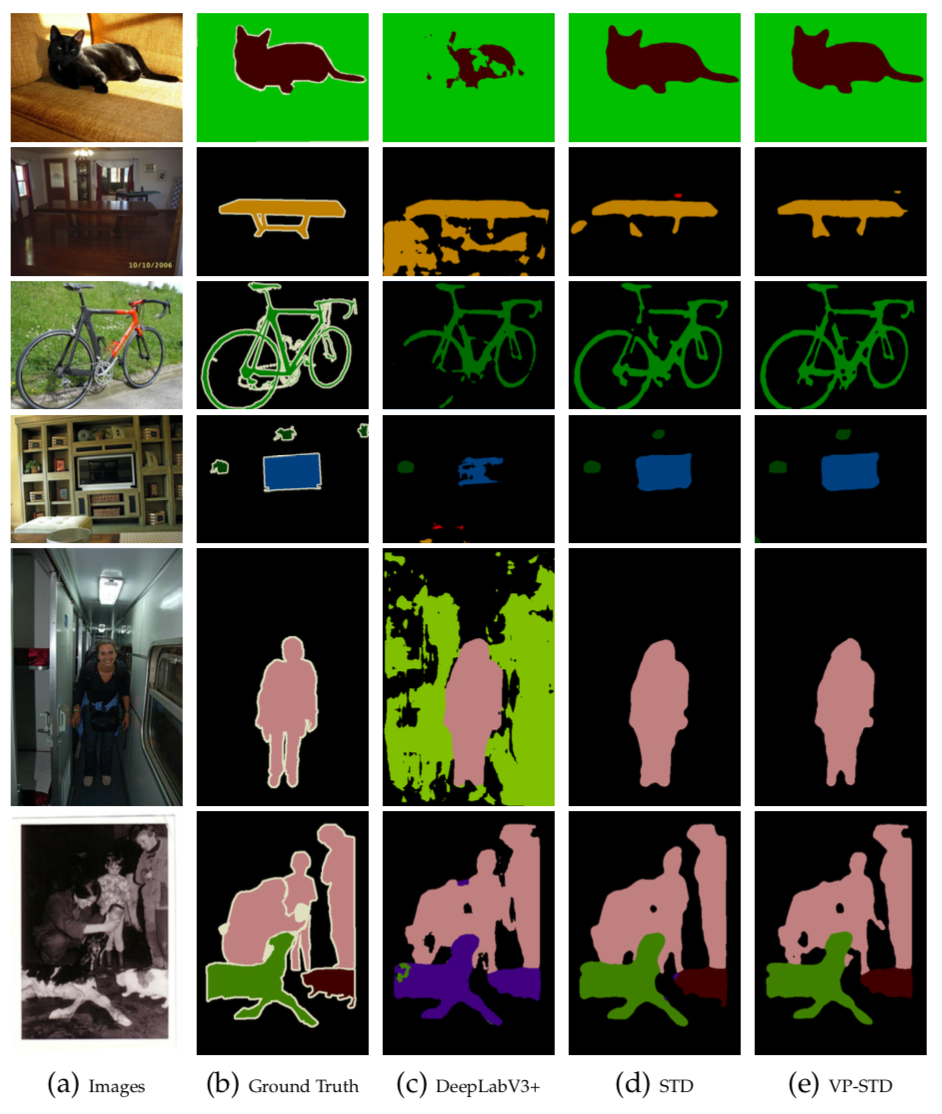}
\caption{Visual effects of the DeepLabV3+, proposed STD-DeeplabV3+, VP-STD-DeeplabV3+ on PASCAL VOC 2012 validation set.}\label{fig6}
\end{figure}

If the segmentation results provided by DeeplabV3+ is near
piecewise constants with smooth boundaries, the proposed method would produce the similar results as DeeplabV3+'s. But when DeeplabV3+ fails to find 
segmentation with spatial priors, the STD based layers can guarantee that the segmentations are near piecewise constants.
In \figurename~\ref{fig6}, parts of the segmentation results
in the validation set produced by DeeplabV3+, STD-DeeplabV3+ and VP-STD-DeeplabV3+ when training are displayed. One can find that the proposed method can produce smooth boundaries for the objects due to the spatial regularization.
Some small wrongly segmented regions by DeepLabV3+ can be removed by our method. In this last figure in \figurename~\ref{fig6}, the misclassification  can be corrected by STD and VP-STD because of the back propagation of regularization and volume information in the network.

\begin{figure}
\includegraphics[width=1.0\linewidth]{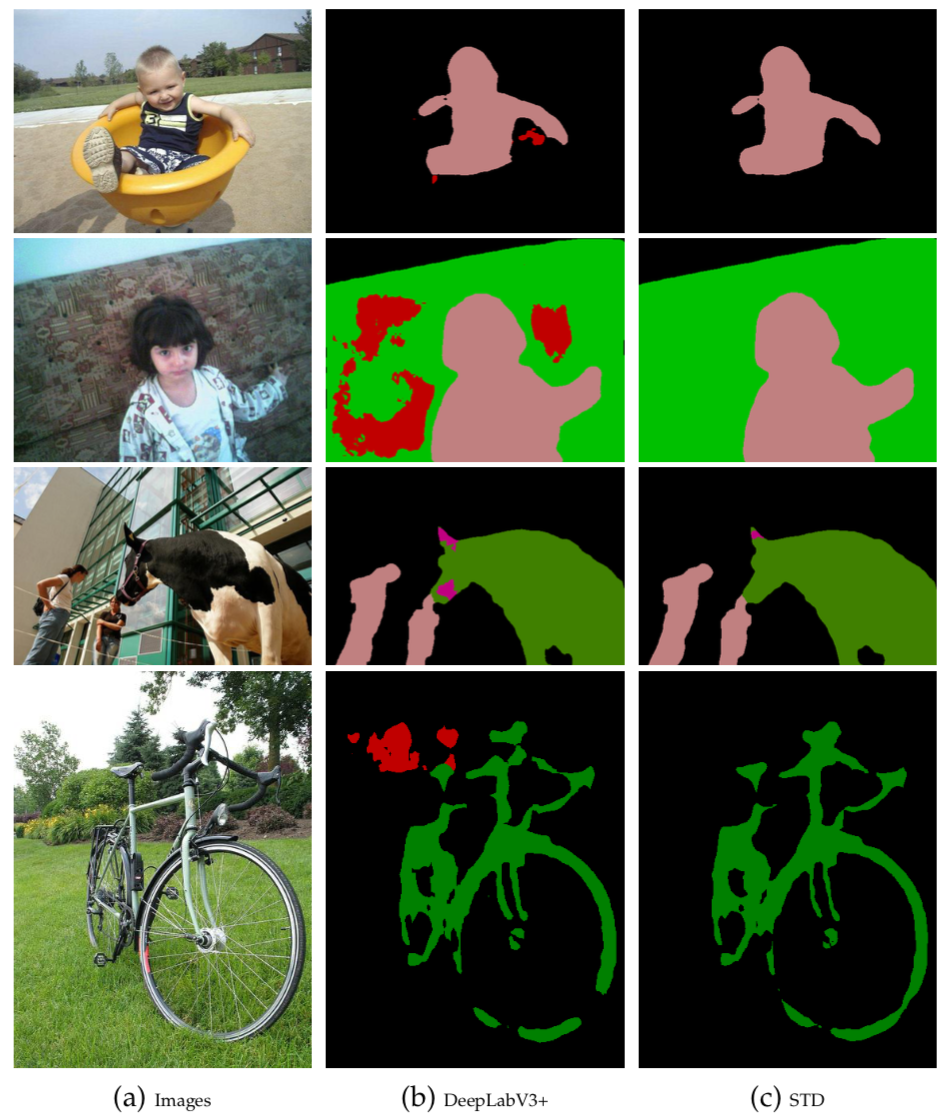}
\caption{Visual effects of the DeepLabV3+\cite{v3+} and proposed STD-DeeplabV3+ on PASCAL VOC 2012 test set.}\label{fig7}
\end{figure}

In the next, we test the performance of the algorithms on the test set (1456 images). The IoU comparisons of  the 21 classes can be found in \tablename~\ref{tab4}. Here we just
use VOC 2012 train \& val set in training for fair comparison.
Our STD-softmax layer can improve mIoU about $1\%$ compared to the original DeepLabV3+. In \figurename~\ref{fig7}, the visual results can be found for 
comparison.

In the current VP-STD, we need to know the volumes of each class, thus we do not apply it on the test set since the volumes are unknown. As mentioned earlier, this can be improved by  the inaccuracy volume constraints methods.

\begin{table}[htp]
\caption{Performance of different methods on PASCAL VOC 2012 validation set. (mIoU: $\%$)}\label{tab3}
\begin{center}
\begin{tabular}{cl|c}
\toprule[2pt]
&Methods&mIoU\\
\midrule[2pt]
\multirow{3}*{Existing}&DeepLabV3+(0 iter.)\cite{v3+}&82.20\\
\cline{2-3}
&DeepLabV3+(50 K iter.)\cite{v3+}&91.34\\
\cline{2-3}
&TV-DeepLabV3+(50 K iter.)\cite{Jia2019}&91.71\\
\cline{2-3}
&VP-TV-DeepLabV3+(50 K iter.)\cite{Li2019}&93.41\\
\midrule[2pt]
\multirow{2}*{Proposed}&STD-DeepLabV3+(50 K iter.)&94.40\\
\cline{2-3}
&VP-STD-DeepLabV3+(50 K iter.)&\textbf{95.21}\\
\bottomrule[2pt]
\end{tabular}
\end{center}
\end{table}

\begin{table*}[htp]
\caption{IoU per class on PASCAL VOC 2012 test set.}\label{tab4}
\begin{center}
\begingroup
\setlength{\tabcolsep}{5pt} 
\renewcommand{\arraystretch}{1.0}
\begin{tabular}{lccccccccccc}
\specialrule{0em}{1pt}{1pt}
\toprule[2pt]
&background&aeroplane&bicycle&bird&boat&bottle&bus&car&cat&chair&cow\\
\cline{2-12}
DeepLabV3+\cite{v3+}&$96.11$&$ \textbf{96.05}$&$62.49$&$ 95.67$&$ 76.45 $&$ 86.41$&$ 96.16$&$ 88.15$&$ \textbf{94.62}$&$ 42.75$&$93.22$\\
Ours (STD)&$\textbf{96.32}$&$ 95.95$&$ \textbf{64.41}$&$ \textbf{95.96}$&$ \textbf{76.46}$&$ \textbf{86.75}$&$ \textbf{96.54}$&$\textbf{ 90.14}$&$ 94.47$&$ \textbf{46.90}$&$\textbf{94.05}$\\
\midrule[2pt]
&diningtable& dog& horse& motorbike&
    person& pottedplant& sheep& sofa& train& tv& mIoU\\
\cline{2-12}
DeepLabV3+\cite{v3+}&$ 81.15$&$\textbf{91.82}$&$96.15$&$92.16$&$90.26$&$64.37$&$88.61$&$61.15$&$82.53$&$\textbf{79.34}$&$83.60$\\
Ours (STD) &$\textbf{81.92}$&$ 91.53$&$ \textbf{96.44}$&$ \textbf{92.57}$&$ \textbf{90.27}$&$ \textbf{65.42}$&$ \textbf{92.15}$&$ \textbf{63.46}$&$ \textbf{83.92}$&$ 79.08$&$\textbf{84.51}$\\
\bottomrule[2pt]
\end{tabular}
\endgroup
\end{center}
\end{table*}

\begin{table}[htp]
\caption{Comparison of training speed ( iteration number /second). }\label{tab5}
\begin{center}
\begin{tabular}{l|c}
\toprule[2pt]
Methods&speed (iter. no. /sec.)\\
\midrule[2pt]
DeepLabV3+\cite{v3+}&1.43\\
\hline
TV-DeepLabV3+\cite{Jia2019}&0.85\\
\hline
VP-TV-DeepLabV3+\cite{Li2019}&0.57\\
\hline
STD-DeepLabV3+&1.21\\
\hline
VP-STD-DeepLabV3+&0.77\\
\bottomrule[2pt]
\end{tabular}
\end{center}
\end{table}

\subsubsection{CITYSCAPE}
\begin{figure*}
\includegraphics[width=1.0\linewidth]{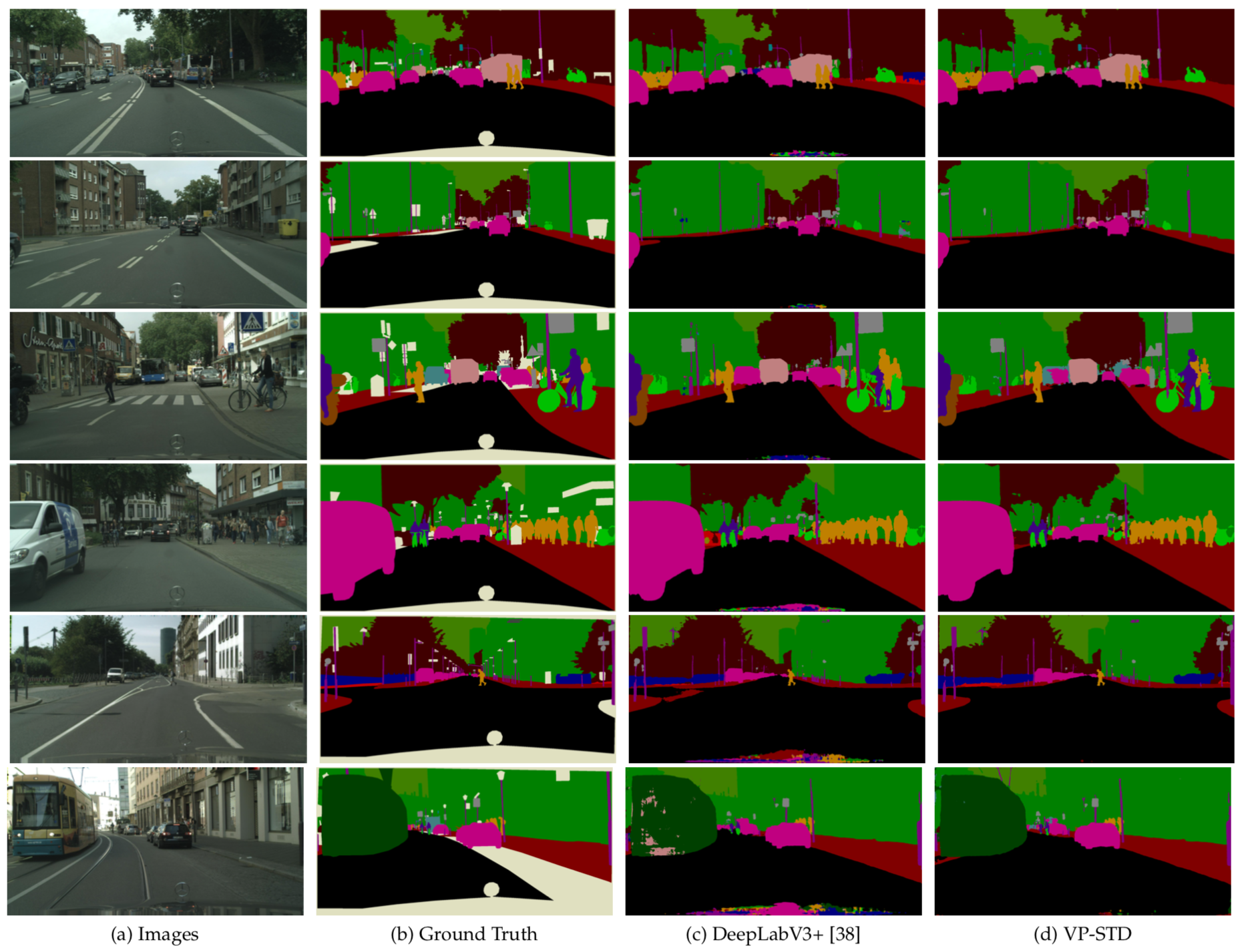}
\caption{Visual effects of the DeepLabV3+,  VP-STD-DeeplabV3+ on CITYSCAPE validation set.}\label{fig8}
\end{figure*}

In this section, the performance of VP-STD layers are evaluated.
There are $2795$ train and $500$ validation images in this dataset \cite{cordts2016cityscapes}. 
The train images are pre-processing with crop size $769\times 769$. For this dataset, the batch size of train 
is $16$. We train the VP-STD based network on the train set using the pre-trained weights ``xception65\_cityscapes\_trainfine" (\url{http://download.tensorflow.org/models/deeplabv3_cityscapes_train_2018_02_06.tar.gz}).
The baseline mIoU given by DeepLabV3+ is $78.73$,
we take the best result of VP-STD-DeepLabV3+
to compare. In the VP-STD block, the $\lambda=1.0,\varepsilon=1.0$ for this dataset, and the iterations of training is set as 90K.
The IoU of 19 classes, the mIoU and visual effects can be found in \tablename~\ref{tab6} and \figurename~\ref{fig8}, respectively. One can see that the mIoU can be improved as $82.07\%$.

\begin{table*}[htp]
\caption{Performance on CITYSCAPE validation set.}\label{tab6}
\begin{center}
\begingroup
\setlength{\tabcolsep}{0.5pt} 
\renewcommand{\arraystretch}{1.5}
\begin{tabular}{l|c|c|c|c|c|c|c|c|c|c|c|c|c|c|c|c|c|c|c|c}
\toprule[2pt]
\hline
&road &s.walk& build.& wall &fence &pole &t-light &t-sign &veg& terrain& sky &person &rider& car& truck& bus& train& motor& bike& mIoU\\
\hline
DeepLabV3+\cite{v3+}&$98.14$&$84.71$&$92.69$&$57.26$&$62.19$&$\bm{65.11}$&$68.41$&$\bm{78.78}$&$92.66$&$63.39$&$95.30$&$82.14$&$62.77$&$\bm{95.31}$&$85.32$&$89.07$&$80.91$&$64.51 $&$77.26$&$78.73$\\
Ours(VP-STD)&$\bm{98.43}$&$\bm{87.68}$&$\bm{93.35}$&$\bm{71.40}$&$\bm{73.58}$&$65.08$&$\bm{68.70}$&$78.38$&$\bm{92.79}$&$\bm{72.59}$&$\bm{95.56}$&$\bm{82.29}$&$\bm{67.43}$&$95.20$&$\bm{89.89}$&$\bm{91.89}$&$\bm{86.56}$&$\bm{70.73}$&$\bm{77.88}$&$\bm{82.07}$\\
\hline
\bottomrule[2pt]
\end{tabular}
\endgroup
\end{center}
\end{table*}

\subsubsection{ISIC 2018}
In this section, the performance of SS-STD softmax layers
are tested. We extracted a data set from the ``ISIC 2018: Skin Lesion Analysis Towards Melanoma Detection" grand challenge datasets (\url{https://challenge.kitware.com/#phase/5abcb19a56357d0139260e53}) \cite{ISIC2018,tschandl2018ham10000}. In this medical images data set, the skin lesions regions and the background need to be segmented. The original dataset contains $2594$ images together with their ground truth in the original dataset. We eliminate the non-star shape object images from the original dataset and select $2198$ images to form a star-shape dataset. All these objects needed to be segmented are almost star-shapes.
We randomly choose $80\%$(1758 images) as training sample and the rest (440 images) for validation set.
The centers $c$ of the star-shape are mainly produced by the geometric center of the ground truth. As a baseline, we take DeeplabV3+ as comparison.  All the images are segmented into 
2 classes. We also take the pre-trained weights ``xception65\_coco\_voc\_trainaug" in VOC 2012 as  initialization. Other parameters are set as the same as the 
PASCAL VOC 2012 dataset. For STD based layers, we set the parameters as $\lambda=0.75, \varepsilon=0.1$.

\begin{figure*}
\includegraphics[width=1.0\linewidth]{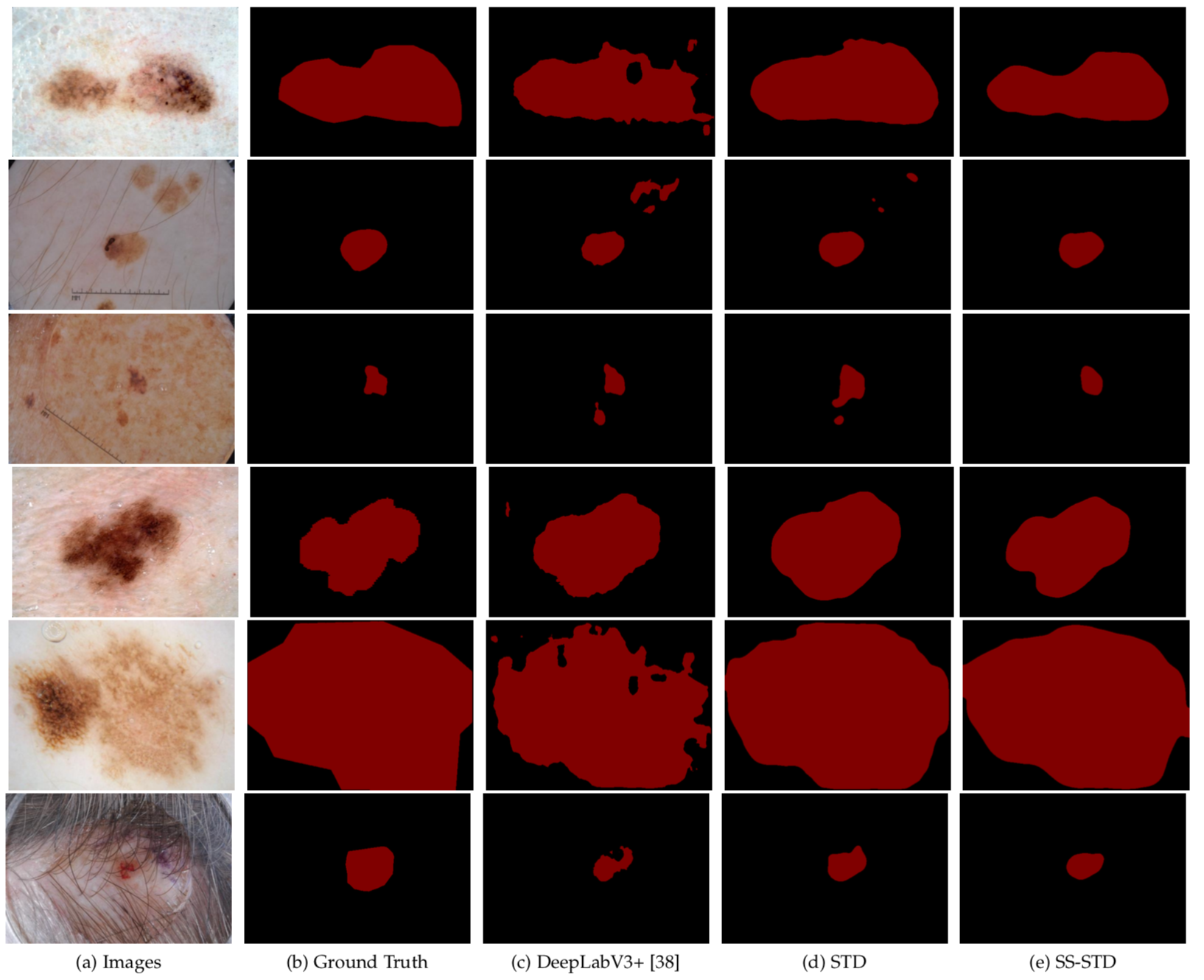}
\caption{Visual effects of the DeepLabV3+, STD-DeeplabV3+, SS-STD-DeeplabV3+ on ISIC2018 validation set.}\label{fig9}
\end{figure*}

The final results can be found in \tablename~\ref{tab7} and \figurename~\ref{fig9}. In this figure, we just list some results which are better than original DeepLabV3+, most of the segmentations are the similar if the predictions are piecewise constants star-shape.

\begin{table}[htp]
\caption{Performance on ISIC2018 validation set.}\label{tab7}
\begin{center}
\begingroup
\setlength{\tabcolsep}{12pt} 
\renewcommand{\arraystretch}{1.0}
\begin{tabular}{ccc}
\toprule[2pt]
&Methods&mIoU\\
\hline
Baseline&DeepLabV3+\cite{v3+}&89.77\\
\hline
\multirow{3}*{Ours}&STD&91.02\\
&VP-STD&92.46\\
&SS-STD&91.57\\
\hline
\bottomrule[2pt]
\end{tabular}
\endgroup
\end{center}
\end{table}

\section{Conclusion and discussion}\label{sec5}
We proposed a STD method to integrate 
the variational priori of the model-based image segmentation into the data-driven DCNN method. In our method, the activation functions appeared in DCNN can be regarded as minimizers of specific variational problems. By finding the dual 
formulation of the variational problem, many important spatial priors such as boundary smoothness, volume preserving, star-shape prior can be incorporated into the DCNN architectures. By the classical model-based segmentation theory, the outputs of the proposed STD based DCNN blocks can have many mathematical properties 
such as smooth boundaries, volume constraints, specific shapes and so on. This is much different from many deep learning works that can not tell the readers why their methods can produce better results. Therefore, if the applications have these priors, our method can greatly improve the accuracy of segmentation. We applied our method to DeepLabV3+ on 
PASCAL VOC 2012, CITYSCAPE and ISIC2018 datasets, experimental results show that the proposed STD-DeepLabV3+ can outperform many state-of-the-art segmentation methods.
Besides, the spatial priori in the proposed method can be
both forward and backward propagated, this is much different from the existing regularization loss function (only backward propagation) and post-processing (only forward propagation) methods.

Our method can be extended to other DCNN based image processing tasks such as image restoration, registration, classification, target detection such as Fast-RCNN \cite{ren2015faster}, Mask-RCNN \cite{he2017mask}.



%

\appendices
\section{Calculating variational of $\mathcal{R}$}\label{appendixA}
Since $\mathcal{R}$ is smooth and thus $\partial \mathcal{R}(\bm u)=\{\delta \mathcal{R}(\bm u)\}$.
Let us first calculate the directional derivative
\begin{equation*}
\begin{array}{rl}
\frac{\mathrm{d} \mathcal{R}(\bm u+\tau \bm v)}{\mathrm{d}\tau}\Big |_{\tau=0}&=\lambda\langle e\bm v, k*(1-\bm u)\rangle-
\lambda\langle e\bm u, k*\bm v\rangle\\
&=\lambda\langle\bm v, (k*(1-\bm u))e-\hat{k}*(e\bm u)\rangle\\
&=\lambda\langle\bm v, (k*(1-\bm u))e-k*(e\bm u)\rangle.
\end{array}
\end{equation*}
Here $\hat{k}$ is the conjugate function of $k$ and the last equation follows by the fact that $\hat{k}=k$  when $k$ is a symmetric kernel function $k(x)=k(-x)$ such as Gaussian kernel.
Therefore
$\delta \mathcal{R}(\bm u)= \lambda ((k*(1-\bm u))e-k*(e\bm u))$ according to the variational equation $\frac{\mathrm{d} \mathcal{R}(\bm u+\tau \bm v)}{\mathrm{d}\tau}\Big |_{\tau=0}=\langle \bm v, \delta \mathcal{R}(\bm u)\rangle$.
\section{Proof of theorem \ref{theo1}}\label{appendixB}

\begin{IEEEproof}
According to \eqref{STDiteration}
\[
\mathcal{F}({\bm{u}}^{t_1+1})+\langle \bm p^{t_1},{\bm{u}}^{t_1+1} \rangle\leq \mathcal{F}({\bm{u}}^{t_1})+\langle \bm p^{t_1},{\bm{u}}^{t_1} \rangle,
\]
and
thus
\begin{equation*}\label{appedixeq1}
\mathcal{F}({\bm{u}}^{t_1+1})-\mathcal{F}({\bm{u}}^{t_1})\leq \langle \bm p^{t_1},{\bm{u}}^{t_1}-{\bm{u}}^{t_1+1} \rangle.
\end{equation*}

Since $\mathcal{R}(\bm{u})$ is concave, by the definition of subgradient for a concave function, $\forall \bm p^{t_1}\in \partial\mathcal{R}(\bm u^{t_1}),$
one can have
\begin{equation*}\label{appedixeq2}
\mathcal{R}({\bm{u}}^{t_1+1})-\mathcal{R}({\bm{u}}^{t_1})\leq \langle \bm p^{t_1},{\bm{u}}^{t_1+1}-{\bm{u}}^{t_1} \rangle.
\end{equation*}
Therefore,
\begin{equation*}
\begin{array}{rl}
&\mathcal{F}({\bm{u}}^{t_1+1})-\mathcal{F}({\bm{u}}^{t_1}) +\mathcal{R}({\bm{u}}^{t_1+1}) -\mathcal{R}({\bm{u}}^{t_1})\\
\leqslant&\langle \bm p^{t_1},{\bm{u}}^{t_1}-{\bm{u}}^{t_1+1} \rangle+\langle \bm p^ t,{\bm{u}}^{t_1+1}-{\bm{u}}^{t_1} \rangle=0,\\
\end{array}
\end{equation*}
which completes the proof.
\end{IEEEproof}

\section{Proof of proposition \ref{pro3}}\label{appendixC}
\begin{IEEEproof}
By introducing Lagrangian multipliers $\bm q, \widehat{\bm q}$ associated to the constraints $\int_{\Omega}u_i(x)\dx=V_i$ and $\sum_{i=1}^I u_{i}(x)=1,\forall x\in\Omega$, 
we have the related Lagrangian functional
\begin{equation*}
\begin{aligned}
&\mathcal{L}(\boldsymbol{u},\bm q,\widehat{\bm q})\\
=&\sum_{i=1}^I \int_{\Omega} (-o_i(x)+p_i^{t_1}(x)+\varepsilon \ln u_i(x))u_i(x)\dx \\
&+\sum\limits^I_{i=1} q_i(V_i-\int_{\Omega}u_i(x)\dx)+\int_{\Omega} \widehat{q}(x)(1-\sum\limits^I_{i=1}u_i(x))\dx.
\end{aligned}
\end{equation*}
Then
\begin{equation*}
\underset{\bm u\in\mathbb{U}_{\bm V}}{\min}\left\{\mathcal{F}(\bm u;\bm o)+\hat{\mathcal{R}}(\bm u)\right\}=\underset{\bm u}{\min}~\underset{\bm q,\widehat{\bm q}}{\max}~\mathcal{L}(\boldsymbol{u},\bm q,\widehat{\bm q}).
\end{equation*}
The derivative of $\mathcal{L}$ with respect to $u_i$
\begin{equation*}\frac{\partial \mathcal{L}}{\partial u_i}=(-o_i+p_i^{t_1}-q_i-\widehat{q}+\varepsilon \ln u_i+\varepsilon=0,
\end{equation*}
therefore, by the first order optimization condition
\begin{equation*}
\widetilde u_i(x)=e^{\frac{o_i(x)-p_i^{t_1}(x)+q_i+\widehat{q}(x)}{\varepsilon}-1},
\end{equation*}
Furthermore, using the condition
\begin{equation*}
\sum\limits^I_{i=1}u_i(x)=e^\frac{\widehat{q}(x)}{\varepsilon}\sum\limits^I_{i=1}e^{\frac{o_i(x)-p_i^{t_1}(x)+q_i}{\varepsilon}-1}=1,
\end{equation*}
we can obtain
\begin{equation*}
\widetilde{ u}_i(x)=\frac{e^{\frac{o_i(x)-q_i^{t_1}(x)+\widetilde{ q}_i(x)}{\varepsilon}}}{\sum\limits^I_{\hat{i}=1}e^{\frac{o_{\hat{i}}(x)-q_{\hat{i}}^{t_1}(x)+\widetilde{q}_{\hat{i}} (x)}{\varepsilon}}},
\label{uvalue}
\end{equation*}
Substituting this into the saddle problem of $\mathcal{L}(\bm u,\bm q,\widehat{\bm q})$, we can obtain
\begin{equation*}
\begin{aligned}
&\mathop{\mathrm{max}}\limits_{\bm q}\left\{\sum\limits^I_{i=1}q_iV_i-\varepsilon\int_{\Omega}\mathrm{\ln}\sum\limits^I_{i=1}e^{\frac{o_i(x)-p_i^{t_1}(x)+q_i}{\varepsilon}}\dx-|\Omega|\varepsilon \right\}\\
=&\mathop{\mathrm{max}}_{\bm q}\left\{\langle\bm q,\bm V\rangle+\langle\bm q^{c,\varepsilon},\bm 1\rangle-|\Omega|\varepsilon\right\},
\end{aligned}
\end{equation*}
which completes the proof.
\end{IEEEproof}

\section*{Acknowledgment}
Liu was supported by the National Key Research and Development Program of China (No. 2017YFA0604903) and the National Natural Science Foundation of China (No. 11871035). The work of Tai was supported by Hong Kong Baptist University through grants RG(R)-RC/17-18/02-MATH, HKBU 12300819 and NSF/RGC grant N-HKBU214-19.

\ifCLASSOPTIONcaptionsoff
  \newpage
\fi



%
\bibliography{IEEEabrv,reference}
\bibliographystyle{IEEEtran}

\end{document}